%% file: AnonymousSubmission2027.tex
\documentclass[letterpaper]{article} 
\usepackage[preprint]{aaai2027}  
\usepackage[hyphens]{url}  
\usepackage{graphicx} 
\usepackage{natbib}  
\usepackage{caption} 
\usepackage{algorithm}
\usepackage{algorithmic}
\usepackage{amsmath}
\usepackage{newfloat}
\usepackage{listings}
\DeclareCaptionStyle{ruled}{labelfont=normalfont,labelsep=colon,strut=off} 
\floatstyle{ruled}
\newfloat{listing}{tb}{lst}{}
\floatname{listing}{Listing}

\usepackage{booktabs}
\usepackage{multirow}

\title{Signature-Guided Capacity Occupancy for Dense Expert Merging}
\author{
    Lingching Tung\textsuperscript{\rm 1},
    Chi-Jui Kim\textsuperscript{\rm 2},
    Beicheng Xu\textsuperscript{\rm 1},
    Yuchen Wang\textsuperscript{\rm 1},
    Bin Cui\textsuperscript{\rm 1}\corresponding
}
\affiliations{
    \textsuperscript{\rm 1}Peking University\\
    \textsuperscript{\rm 2}Independent Researcher \\
}

\newcommand{\method}{\textsc{SigMerge}}

\begin{document}

\maketitle

\begin{abstract}
Dense expert merging combines domain-specialized language models into one single checkpoint, typically by admitting task-vector support in weight space. 
However, this admission is governed by three decisions that existing methods answer only partially: where to open layer capacity from cross-expert conflict, who should occupy that capacity based on domain demand, and how to admit the resulting support without relying on costly recipe search.
To tackle these issues, we propose \method{} (\emph{Signature-Guided Capacity Occupancy}), a structured capacity assignment framework for dense expert merging. Starting from a dense base merge, conflict signatures set each layer's capacity from cross-expert conflict, positive base-merge deficits set each domain's share of that capacity, and a sequential occupancy rule admits each expert delta up to the resulting layer-domain budget.
Across 21 paired settings spanning seven dense base merges and three model pools, \method{} improves every one (by 15.0\% on average) and achieves the best average rank (1.67) among six merging methods, outperforming three categories of merging baselines.

\end{abstract}


\section{Introduction}
\label{sec:intro}

Model merging avoids the cost of deploying multiple fine-tuned checkpoints by combining them into a single model of the same size, through arithmetic on their weights and with no further training~\citep{wortsman2022model, ilharco2023editing}.
In this setting, each expert is represented by its task vector, and a merge operator collapses these vectors into one parameter set.
The difficulty is that a single set of shared parameters must encode skills that were acquired in isolation: accommodating the update one expert requires can erase what another has learned.
Most existing methods address this in one shot, applying a single merge operator to the reference checkpoint and the task vectors and treating its output as final~\citep{yadav2023ties, yu2024language}.
The operator must therefore serve two functions: forming a coarse consensus across experts and recovering the specialist behavior that this consensus degrades.
The second function, however, has received comparatively little attention.
This leaves a basic empirical question: how much specialist capability does a dense merge actually give up?

\begin{figure}[t]
\centering
\includegraphics[width=\columnwidth]{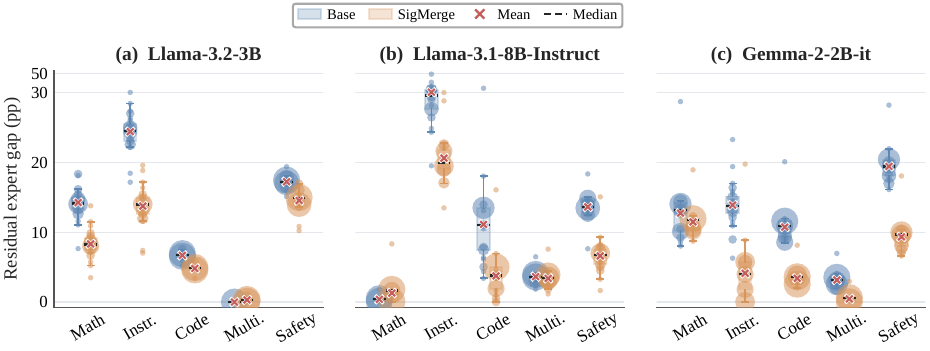}
\caption{Residual expert gap on three model pools and five domains. Blue denotes the dense base merge and orange denotes \method{} applied on top. Lower is better.}
\label{fig:motivation}
\end{figure}

Figure~\ref{fig:motivation} provides the empirical starting point.
The residual expert gap measures how far a merged checkpoint falls short of the best expert on a domain. 
Across seven merge operators spanning two families and three MergeBench model pools~\citep{he2025mergebench}, dense merges leave an average gap of 12.13 points, and these deficits vary substantially across expert pools, domains, and merge families.
This variation exposes a fundamental limitation of uniform merging, since a rule that works well on average may still be poorly suited to particular domains or parts of the model.
Domain-level gaps reveal which capabilities are not being fully retained, but they do not show where those capabilities are lost.
Translating these deficits into effective merge decisions therefore requires evidence at the layer level.

One way to obtain that evidence is to open layer-wise freedom everywhere and let a search supply it.
A convenient proxy, such as negative log-likelihood on a small calibration set, can guide the search, but optimizing this proxy may harm specialist capabilities that it does not capture.
This creates a fundamental trade-off.
Giving every layer the same amount of freedom can be under-adaptive, whereas letting a search set that amount layer by layer can be over-adaptive.
A better approach is to bound the capacity of each layer by limiting how many coordinates the experts may influence.
Importantly, this capacity should be determined from directly measurable signals rather than by searching over merge configurations.

Once the capacity is defined, allocating it introduces three challenges, each of which must be resolved without training.
\textbf{C1.} Deciding where and how many coordinates a layer may give away requires measuring cross-expert disagreement.
\textbf{C2.} Deciding which domains receive those coordinates requires measuring under-retained capabilities that task-vector geometry alone does not reveal.
\textbf{C3.} Admitting them means filling a budget table with one entry per layer and domain, whose entries are costly to search and whose domains compete for the same high-magnitude coordinates.

In this paper we propose \method{}, a training-free operator that recovers the specialist ability a dense merge leaves behind by allocating a bounded number of task-vector coordinates across layers and domains. 
The contributions are summarized as follows.
1) We show that dense merging leaves specialist deficits that are large and heterogeneous across models, domains, and merge families, and that repairing them requires allocating coordinates rather than choosing a better averaging rule.
2) For C1, \method{} folds several views of cross-expert disagreement into one layer-wise conflict signature and maps it through a leaky policy to how many coordinates each layer may give away.
3) For C2, \method{} measures the positive gap between each domain expert and the merged model and divides the opened capacity in proportion to those deficits.
4) For C3, \method{} fills the resulting layer-domain budgets with a sequential top-fraction rule that skips coordinates already claimed by earlier domains.
Empirically, \method{} lowers the average residual expert gap in Figure~\ref{fig:motivation} from 12.13 to 5.77 points.
Applied on top of seven dense merges across three model pools, it improves Macro-5 by 15.0\% relative, and the resulting checkpoints outperform baselines across three categories, attaining the best average rank of 1.67 among six merging methods.

\section{Background}
\label{sec:background}

\subsection{Dense Merging}
\label{sec:bg_problem}
We are given a shared reference checkpoint $\theta_{\mathrm{ref}}$ and a pool of domain experts $M=\{\theta_1,\ldots,\theta_{|\mathcal{D}|}\}$, each fine-tuned from $\theta_{\mathrm{ref}}$ on a distinct domain $d\in\mathcal{D}$. Following task-vector formulations~\citep{ilharco2023editing,wortsman2022model}, expert $d$ is represented by its task vector $\Delta_d=\theta_d-\theta_{\mathrm{ref}}$. Dense expert merging seeks a single dense checkpoint $\widehat{\theta}=\theta_{\mathrm{ref}}+\sum_{d\in\mathcal{D}} U_d$ that maximizes aggregate multi-domain performance:
\begin{equation}
\max_{\{U_d\}}\ \frac{1}{|\mathcal{D}|}\sum_{d\in\mathcal{D}} Q_d\!\Big(\theta_{\mathrm{ref}}+\textstyle\sum_{d'\in\mathcal{D}} U_{d'}\Big),
\label{eq:bg_objective}
\end{equation}
where $Q_d$ is an evaluation score on domain $d$ and each admitted update $U_d$ rescales $\Delta_d$ coordinate-wise, so that $\mathrm{supp}(U_d)\subseteq\mathrm{supp}(\Delta_d)$. Existing operators instantiate this in a single shot, mapping $\theta_{\mathrm{ref}}$ and the task vectors directly to a final checkpoint.

Choosing the support $\{U_d\}$ requires answering three questions: (1)~\emph{where}: in which layers support may open, and how much; (2)~\emph{who}: which domains should occupy that support; and (3)~\emph{how}: how that support is admitted into the dense checkpoint.

\subsection{Merge Strategies}
\label{sec:bg_strategies}

Existing strategies answer these three questions only partially (Table~\ref{tab:bg_strategies}). \emph{Weight-space operators} (Model Soups~\citep{wortsman2022model}, Task Arithmetic~\citep{ilharco2023editing}, Fisher merging~\citep{matena2022merging}, RegMean~\citep{jin2023dataless}) apply one fixed, domain-agnostic rule to every layer, with no conflict resolution. \emph{Conflict-aware methods} (TIES~\citep{yadav2023ties}, DARE~\citep{yu2024language}, CABS~\citep{yang2025cabs}, CAT~\citep{sun2025cat}, and localized variants~\citep{wang2024localizing,yan2025calm}) control admitted delta mass per coordinate, component, or support. \emph{Layer-wise methods} (LiNeS~\citep{wang2025lines}, Layer Swapping~\citep{bandarkar2025layer}, LOT~\citep{sun2025towards}, AdaMerging~\citep{yang2024adamerging}) vary the merge with depth, from depth priors or learned objectives. \emph{Search-based methods} (evolutionary optimization~\citep{akiba2025evolutionary}, MERGE$^3$~\citep{mencattini2025merge}, Pareto Merging~\citep{chen2025pareto}) search the full recipe, relying on proxies that only approximate true performance. 

\begin{table}[t]
\centering
{\small
\setlength{\tabcolsep}{5pt}
\begin{tabular}{lccc}
\hline
Family & \emph{Where} & \emph{Who} & \emph{How} \\
\hline
Weight-space   & No & No & Yes \\
Conflict-aware & No & No & Yes \\
Layer-wise     & Partial & No & Partial  \\
Search-based   & Partial & Partial & No  \\
\hline
\method{} (ours) & Yes & Yes & Yes \\
\hline
\end{tabular}}
\caption{How existing families and \method{} cover the three questions: \emph{where}, \emph{who}, and \emph{how}.}
\label{tab:bg_strategies}
\end{table}

These partial answers say which questions go unanswered, but not why. Each gap traces to a distinct limitation of current designs, which we state below as one issue per question.

\textbf{Issue \#1: Layer capacity is not grounded in conflict.} Conflict-aware methods measure cross-expert conflict but consume it per coordinate, while layer-wise methods vary admission with depth from priors or learned objectives; neither sets a layer's capacity from the conflict present in it.

\textbf{Issue \#2: Domain demand is ignored.} Admission is decided per coordinate, component, or support, not by which domains the merge still under-serves, so the opened capacity is not routed to the domains that need it most.

\textbf{Issue \#3: Layer and domain specific admission requires search.} Beyond per-coordinate sparsification, the only route is to search the full recipe, which exposes $L\times|\mathcal{D}|$ choices and is sensitive to weak proxies.
It also cannot prevent multiple domains from claiming the same coordinates, reducing distinct support under a fixed budget.

\section{Method}
\label{sec:method}

\begin{figure*}[t]
\centering
\includegraphics[width=0.8\linewidth]{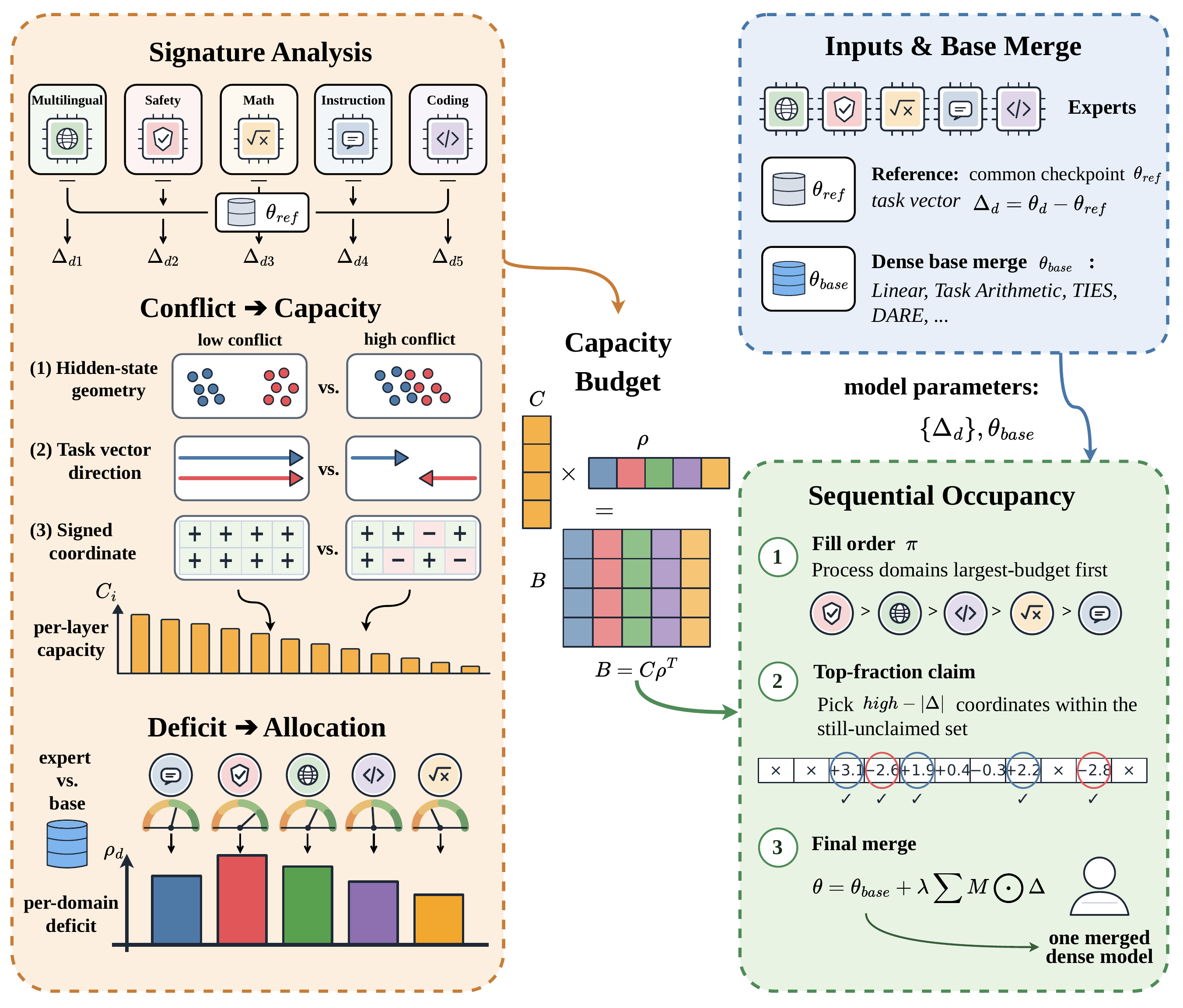}
\caption{Overview of \method{}.}
\label{fig:overview}
\end{figure*}

\method{} admits a bounded amount of extra expert-delta support around a dense base merge. How much a layer may expose is read from how far its experts disagree, and how that is divided among domains from what the base merge failed to keep. Their product is a layer-domain budget that a claiming rule turns into coordinates. Fig.~\ref{fig:overview} shows the construction.

\subsection{Capacity Budgets}
\label{sec:capacity_budgets}

Eq.~\ref{eq:bg_objective} leaves $\{U_d\}$ free. A dense base merge is one choice of $\{U_d\}$, which we write $U_d^{0}$, so that $\theta_{\mathrm{base}}=\theta_{\mathrm{ref}}+\sum_{d}U_d^{0}$, where every sum over $d$ runs over $\mathcal{D}$. \method{} keeps that choice and adds a bounded increment $V_d$,
\begin{equation}
\widehat{\theta}=\theta_{\mathrm{ref}}+\sum_{d}(U_d^{0}+V_d)=\theta_{\mathrm{base}}+\sum_{d}V_d ,
\label{eq:sg_form}
\end{equation}
which remains of the form Eq.~\ref{eq:bg_objective} requires. What follows specifies $V_d$. We decompose each checkpoint into $L$ layer blocks $\theta=(\theta_0,\ldots,\theta_{L-1})$, where block $i$ has flattened coordinate set $\Omega_i$ and layer-wise task vector $\Delta_{d,i}=\theta_{d,i}-\theta_{\mathrm{ref},i}$.

Two things about $V_{d,i}$ could be limited, the magnitude it writes into a layer or how many coordinates it writes to. We limit the second, for two reasons. The magnitude of $\Delta_{d,i}$ is what the expert actually learned, so rescaling it per layer distorts that signal while saying nothing about how the layer should be divided. The count, in contrast, is precisely what the experts contend over, because a coordinate handed to one domain is a coordinate the others can no longer use. Occupancy, the fraction $|\mathrm{supp}(V_{d,i})|/|\Omega_i|$, is therefore the quantity a budget can meaningfully cap, and we write $B_{i,d}\in[0,1]$ for that cap.

Every entry of $B$ has to come from some measurement, and what we can measure fixes what shapes $B$ can take. Without training, the evidence available to us is of two kinds and each comes at its own granularity. How contested a layer is can be read from the experts at that layer, giving one number per layer. How much a domain still lacks can be read from that domain's score, giving one number per domain. Neither carries information about a particular layer-domain pair. Filling $L|\mathcal{D}|$ cells individually would therefore require inventing the joint part, either by searching it against a proxy or by positing an interaction we cannot measure. Two such sources can support an outer product and nothing finer, so we set
\begin{equation}
B=C\rho^{\top},\qquad B_{i,d}=C_i\rho_d ,
\label{eq:budget}
\end{equation}
with conflict shaping only $C$ and deficit shaping only $\rho$, using a per-layer capacity $C_i\in[C_{\min},C_{\max}]$ and a per-domain profile $\rho$ on the simplex over $\mathcal{D}$, allowing also $\rho=\mathbf{0}$ for the case where no domain is under-retained. The table is then described by $L+|\mathcal{D}|$ controls rather than $L|\mathcal{D}|$ free entries, and $\sum_{d}B_{i,d}=C_i$ whenever $\rho$ sums to one, so $C_i$ is the total support layer $i$ exposes. The remaining work is to determine the two factors and then to realize their product as coordinates.

\subsection{Conflict to Capacity}
\label{sec:conflict_capacity}

The per-layer capacity $C$ decides how much of each layer may be occupied.
A layer whose experts already agree can take on expert-specific support without one expert undoing another, so what $C$ needs is a per-layer reading of how far they disagree.
Measuring this agreement, however, requires more than one statistic.
Experts can move a layer in nearly the same direction and still demand opposite signs on individual weights, and they can move it in near-orthogonal directions and still leave its function unchanged, which only a reading of what the layer computes can detect. 
Since each statistic captures only part of this behavior, and no held-out objective tells us which one to prioritize, we measure disagreement from three complementary views: the layer's representations, the directions of its expert updates, and their coordinate-wise signs,
\begin{equation}
\begin{array}{r@{\;}c@{\;}l}
r_i &=& \mathrm{avg}_{\mathcal{P}}\big(1-\mathrm{CKA}(H_{d,i},H_{d',i})\big), \\[2pt]
p_i &=& \mathrm{avg}_{\mathcal{P}}\big( \frac{1-\cos(\Delta_{d,i},\Delta_{d',i})}{2}\big), \\[2pt]
s_i &=& \displaystyle\frac{\sum_{j} m_{i,j}\,c_{i,j}}{\sum_{j} m_{i,j}+\varepsilon},
\end{array}
\label{eq:views}
\end{equation}
where $\mathrm{avg}_{\mathcal{P}}$ averages over the set $\mathcal{P}$ of expert pairs, $\mathrm{CKA}$ is the centered kernel alignment~\citep{kornblith2019similarity}, $H_{d,i}$ are pooled hidden states on a shared set of probe prompts, $j$ ranges over $\Omega_i$, $m_{i,j}=\max_{d}|\Delta_{d,i}[j]|$ is the largest demand on coordinate $j$, and $c_{i,j}$ marks the coordinates where some pair demands opposite signs. Larger is more contested in all three.

The three occupy different ranges across layers, so we min-max each of them over layers and take the unweighted mean $\bar{S}_i\in[0,1]$. Weighting them would require deciding which reading counts more, and the reason for taking three is that we cannot decide that without a held-out objective. Capacity is linear in that score,
\begin{equation}
C_i=C_{\max}-(C_{\max}-C_{\min})\,\bar{S}_i ,
\label{eq:capacity}
\end{equation}
running from $C_{\max}$ at $\bar{S}_i=0$ down to a floor $C_{\min}>0$ at $\bar{S}_i=1$. The floor is there because $\bar{S}_i$ is relative. A top-scoring layer is the most contested among these layers, not contested in absolute terms, and closing it entirely would withdraw support on that evidence alone. 

\subsection{Deficit to Allocation}
\label{sec:deficit_allocation}

$C$ says how much of each layer is available but not who receives it, and the three readings cannot say either. They are taken from weights and activations, which describe how the experts differ from one another, not which of their capabilities the base merge kept. That is a question about outcomes, and outcomes are what $Q_d$ measures. For a checkpoint $\theta$, define the residual expert gap on domain $d$ as
\begin{equation}
g_d(\theta)=\max\big(0,\ Q_d(\theta_d)-Q_d(\theta)\big),
\label{eq:gap}
\end{equation}
the amount by which $\theta$ falls short of that domain's own expert. Taking that expert as the reference keeps the target inside what the pool already contains, so the deficit measures what merging lost rather than what the domain could reach.

The allocation reads this gap at the base merge, scoring on a calibration split $\mathcal{C}_d$ taken from the domain's benchmark and holding out the remainder. The same $g_d$ serves as the reported metric on that remainder, which is what Fig.~\ref{fig:motivation} plots.

Capacity is divided across domains in proportion to their deficits at the base merge,
\begin{equation}
\rho_d=\frac{g_d(\theta_{\mathrm{base}})}{\sum_{d'}g_{d'}(\theta_{\mathrm{base}})} ,
\label{eq:share}
\end{equation}
and no capacity is opened at all when every domain already matches its expert. The clip at zero in Eq.~\ref{eq:gap} is what produces that behavior, and more generally it is what keeps the opened capacity aimed only at what is still missing. Because of Eq.~\ref{eq:budget}, one $\rho$ applies at every layer, so a domain's share of the opened capacity does not change with depth and only the total opened does. That follows from where the two signals live. Conflict is measured layer by layer and can vary with depth, whereas a deficit is a property of the merged model as a whole and says nothing about which layers are responsible for it.

\begin{algorithm}[t]
\caption{\method{}}
\label{alg:sigmerge}
\begin{algorithmic}[1]
\REQUIRE $\theta_{\mathrm{ref}}$, experts $\{\theta_d\}$, base merge $\theta_{\mathrm{base}}$, calibration splits $\{\mathcal{C}_d\}$, constants $C_{\min},C_{\max},\lambda$
\ENSURE merged checkpoint $\widehat{\theta}$
\STATE $\Delta_d \leftarrow \theta_d-\theta_{\mathrm{ref}}$ for all $d$
\STATE $\tilde{r},\tilde{p},\tilde{s} \leftarrow$ min-max of $r,p,s$ across layers \COMMENT{Eq.~\ref{eq:views}}
\STATE $\bar{S}_i \leftarrow \mathrm{mean}(\tilde{r}_i,\tilde{p}_i,\tilde{s}_i)$ for all $i$
\STATE $C_i \leftarrow C_{\max}-(C_{\max}-C_{\min})\bar{S}_i$ for all $i$ \COMMENT{Eq.~\ref{eq:capacity}}
\STATE $G \leftarrow \sum_{d} g_d(\theta_{\mathrm{base}})$ \COMMENT{Eq.~\ref{eq:gap}, scored on $\mathcal{C}_d$}
\IF{$G=0$}
  \RETURN $\theta_{\mathrm{base}}$
\ENDIF
\STATE $\rho_d \leftarrow g_d(\theta_{\mathrm{base}})/G$ for all $d$ \COMMENT{Eq.~\ref{eq:share}}
\STATE $\widehat{\theta} \leftarrow \theta_{\mathrm{base}}$
\FOR{each layer $i$}
  \STATE $\mathcal{A} \leftarrow \Omega_i$
  \FOR{$d$ in decreasing order of $\rho_d$}
    \STATE $M \leftarrow$ the $\lceil C_i\rho_d|\Omega_i|\rceil$ entries of $\mathcal{A}$ with largest $|\Delta_{d,i}|$ \COMMENT{Eq.~\ref{eq:budget}}
    \STATE $\widehat{\theta}_i \leftarrow \widehat{\theta}_i + \lambda\,\Delta_{d,i}$ restricted to $M$
    \STATE $\mathcal{A} \leftarrow \mathcal{A} \setminus M$
  \ENDFOR
\ENDFOR
\RETURN $\widehat{\theta}$
\end{algorithmic}
\end{algorithm}

\subsection{Sequential Occupancy}
\label{sec:sequential}

The budget specifies how many coordinates each domain may use, but not which ones.
This choice creates competition because different domains often assign their largest updates to the same coordinates.
If each domain selects from the full layer independently, their supports overlap.
Repeatedly selecting the same coordinate does not increase the distinct support used, leaving part of the available capacity unfilled.

To avoid this, domains select coordinates sequentially, as shown in Alg.~\ref{alg:sigmerge}.
Each domain chooses the largest-magnitude entries of its task vector from the coordinates that remain available.
Since $C_i$ is common to every domain in a layer, ordering by budget is ordering by deficit, and the order is the same at every layer. Serving the largest deficit first matters because each later domain chooses from a smaller set, and the domain furthest behind is the one least able to afford that restriction. Selecting by $|\Delta_{d,i}|$ within a domain spends the budget where the expert moved furthest from the reference, which is where dropping it costs that domain most. Writing $M_{i,d}$ for the resulting mask, the admitted increment is $V_{d,i}=\lambda\,M_{i,d}\odot\Delta_{d,i}$ and the layer becomes
\begin{equation}
\widehat{\theta}_i=\theta_{\mathrm{base},i}+\lambda\sum_{d} M_{i,d}\odot\Delta_{d,i} .
\label{eq:occupancy}
\end{equation}

The same scale $\lambda$ is applied to every selected update, regardless of layer or domain.
The procedure requires no search. It reads each signal once using one forward pass over the probe prompts, one pass over the task vectors, and one score per domain.

\section{Experiments}
\label{sec:experiments}


\subsection{Experimental Setup}
\label{sec:exp_setup}

\paragraph{Model pools and experts.}
The three MergeBench pools~\citep{he2025mergebench} are Llama-3.2-3B, Llama-3.1-8B-Instruct, and Gemma-2-2B-it, and we name them in that order throughout. Each pool supplies five experts fine-tuned from a shared reference checkpoint, one for each of math, instruction, coding, multilingual, and safety.

\paragraph{Anchors.}
We apply \method{} on top of seven dense merges, and we write \emph{anchor} for the dense merge a checkpoint is built on. All seven come from the two families of Table~\ref{tab:bg_strategies} that map the task vectors to a single checkpoint. Four are weight-space operators, 1)~Linear expert soup~\citep{wortsman2022model}, 2)~Task Arithmetic (TA)~\citep{ilharco2023editing}, 3)~Fisher merging~\citep{matena2022merging}, and 4)~RegMean~\citep{jin2023dataless}. Three are conflict-aware operators, 5)~TIES~\citep{yadav2023ties}, 6)~DARE~\citep{yu2024language}, and 7)~DARE-TIES~\citep{yadav2023ties,yu2024language}. Crossed with the three pools this gives 21 anchor-pool settings.

\paragraph{Baselines.}
We compare the resulting checkpoints against five further methods. Two carry the conflict-aware family past the anchors above. 1)~CABS~\citep{yang2025cabs} sparsifies each task vector by $n{:}m$ block-wise magnitude pruning under sequential masks that keep the retained coordinates disjoint. 2)~CAT Merging~\citep{sun2025cat} trims conflict-prone components layer by layer, projecting linear weights and masking normalization parameters from a few unlabeled exemplars. The remaining three cover the two families of Table~\ref{tab:bg_strategies} that no anchor represents. 3)~LiNeS~\citep{wang2025lines} rescales the task vector with a coefficient that grows with depth, shrinking shallow-layer updates. 4)~LOT Merging~\citep{sun2025towards} solves a closed-form layer-wise least-squares problem on exemplar features to minimize feature drift from each expert. 5)~MERGE$^3$~\citep{mencattini2025merge} evolves merge configurations against an item-response-theory estimate computed on a reduced evaluation subset. All five run on the same expert pools, benchmark suite.

\paragraph{Metrics.}
The primary metric is Macro-5, the unweighted mean of the five domain aggregates under the MergeBench~\citep{he2025mergebench} grouping, computed from GSM8K~\citep{cobbe2021trainingverifierssolvemath} (math), IFEval~\citep{zhou2023instructionfollowingevaluationlargelanguage} (instruction), HumanEval+/MBPP+~\citep{chen2021evaluatinglargelanguagemodels,austin2021programsynthesislargelanguage,liu2023codegeneratedchatgptreally} (coding), the Okapi multilingual versions of mMMLU/mARC/mHellaSwag~\citep{lai2023okapiinstructiontunedlargelanguage,hendrycks2021measuringmassivemultitasklanguage,clark2018thinksolvedquestionanswering,zellers2019hellaswagmachinereallyfinish} (multilingual), and WildGuardTest/HarmBench/XSTest/DAN~\citep{han2024wildguardopenonestopmoderation,mazeika2024harmbenchstandardizedevaluationframework,röttger2024xstesttestsuiteidentifying,shen2024donowcharacterizingevaluating} (safety). We also report the residual expert gap $g_d$ of Eq.~\ref{eq:gap}, taking the reference on domain $d$ to be the strongest expert in the pool on that domain. Submetric breakdowns are in the supplementary material.

\paragraph{Implementation details.}
We fix $C_{\min}=0.10$, $C_{\max}=0.45$, and $\lambda=0.60$ for every pool and every anchor, and these three constants produce every number below. Deficits are estimated on a calibration split of each domain's benchmark, and all scores are reported on the disjoint remainder. All merging and evaluation runs use a single NVIDIA H100 80GB GPU under Python~3.10. DARE and DARE-TIES draw random masks, so each is run with three seeds, giving 11 anchor runs per pool and 33 in total. MERGE$^3$ searches stochastically and is likewise run with three seeds. All tables report seed~0. Per-seed statistics, sweeps over the three constants, the full software stack, and the cost of obtaining each checkpoint are in the supplementary material.

\subsection{Effectiveness of Deficit-Driven Allocation}
\label{sec:exp_deficit}

\begin{figure}[t]
\centering
\includegraphics[width=\linewidth]{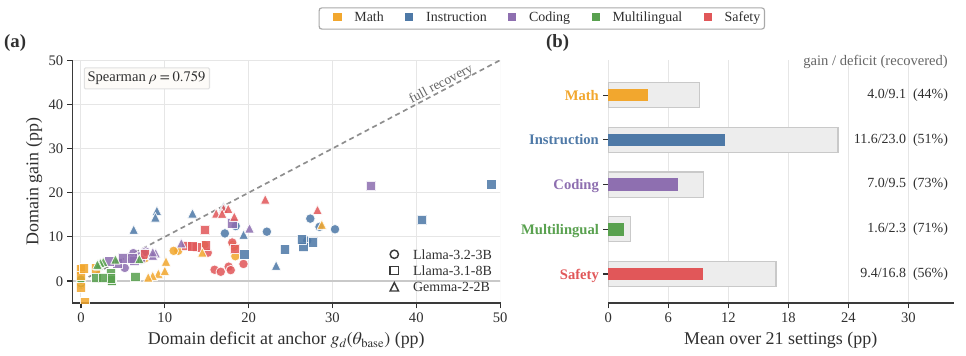}
\caption{Deficit-driven allocation. (a)~Gain on a domain against that domain's residual expert gap at the anchor, one point per (pool, anchor, domain) triple. Marker encodes the pool. Color encodes the domain and is shared with panel~(b). The dashed line marks full recovery of the deficit. (b)~Deficit at the anchor and gain from \method{}, averaged over the 21 anchor-pool settings. Labels give the recovered fraction.}
\label{fig:deficit_gain}
\end{figure}

\begin{table}[t]
\centering
{\small
\setlength{\tabcolsep}{4pt}
\begin{tabular}{lccc}
\hline
Pool & anchor & $+$\method{} & reduction \\
\hline
Llama-3.2-3B          & 12.52 & 6.92 & $-5.60$ \\
Llama-3.1-8B-Instruct & 11.88 & 6.46 & $-5.42$ \\
Gemma-2-2B-it         & 12.00 & 3.93 & $-8.07$ \\
\hline
All                   & \textbf{12.13} & \textbf{5.77} & $\mathbf{-6.36}$ \\
\hline
\end{tabular}}
\caption{Residual expert gap $g_d$ averaged over the five domains and the seven anchors of each pool, in percentage points. Lower is better.}
\label{tab:residual_gap}
\end{table}

Fig.~\ref{fig:deficit_gain} tests \textbf{C2}, the claim that domain shares should follow measured deficits (Sec.~\ref{sec:deficit_allocation}). Averaged over the three pools, \method{} cuts the residual expert gap from 12.13 to 5.77 points (Table~\ref{tab:residual_gap}). Where that reduction comes from is visible once the gap is broken out by domain. Fig.~\ref{fig:deficit_gain} plots the gain on each domain against that domain's deficit at the anchor, for all 105 (pool, anchor, domain) triples. The three pools do not always agree, which is the heterogeneity Sec.~\ref{sec:intro} reports, so we split the aggregate wherever they diverge. Three observations follow. 1)~\emph{The allocation tracks the deficit.} The rank correlation between deficit at the anchor and realized gain is Spearman $+0.759$ over the 105 triples, and $+0.77$, $+0.85$, and $+0.61$ within the three pools. Since $\rho_d$ is the only factor of Eq.~\ref{eq:budget} that varies across domains, the ordering of gains is the ordering Eq.~\ref{eq:gap} writes into $\rho$. 2)~\emph{Multilingual behaves as a control group.} Multilingual carries the smallest deficit, 2.27 points on average, an order of magnitude below instruction. That average covers three different regimes. On Llama-3.2-3B all seven anchors already match the multilingual expert, so $\rho$ is exactly zero and the checkpoint moves by $-0.07$. On the other two pools the deficit is 3.64 and 3.17 points and \method{} adds 0.68 and 4.20. The clip at zero in Eq.~\ref{eq:gap} is what produces the first case. A domain that is not behind registers no demand and is allocated nothing at all. 3)~\emph{No domain is fully recovered.} Instruction and safety carry the two largest deficits, 22.97 and 16.77 points, and \method{} recovers 50.7\% and 56.3\% of them. The rate does not follow the size of the deficit. Coding and math start from almost the same deficit, 9.52 and 9.13 points, and recover 73.5\% and 43.9\%, and on Gemma-2-2B-it instruction and math start from 13.89 and 12.74 points and recover 90.7\% and 33.4\%.

\subsection{Effectiveness of the Capacity Decomposition}
\label{sec:exp_decomposition}

\begin{table}[t]
\centering
{\small
\setlength{\tabcolsep}{4pt}
\begin{tabular}{lcc}
\hline
Variant & Macro-5 & $\Delta$ \\
\hline
\textbf{\emph{Llama-3.2-3B, DARE anchor}}&& \\
default                   & 46.38 & \phantom{$-$}0.00 \\
uniform capacity          & 39.92 & $-6.46$ \\
equal domain share        & 41.29 & $-5.09$ \\
sparse-rescaled update    & 44.46 & $-1.92$ \\
\hline
\textbf{\emph{Llama-3.1-8B-Instruct, Linear anchor}}&& \\
default                   & 65.51 & \phantom{$-$}0.00 \\
uniform capacity          & 60.19 & $-5.32$ \\
equal domain share        & 59.65 & $-5.86$ \\
sparse-rescaled update    & 66.03 & $+0.52$ \\
\hline
\textbf{\emph{Gemma-2-2B-it, Linear anchor}}&& \\
default                   & 56.98 & \phantom{$-$}0.00 \\
uniform capacity          & 46.60 & $-10.38$ \\
equal domain share        & 46.06 & $-10.92$ \\
sparse-rescaled update    & 44.80 & $-12.18$ \\
\hline
\end{tabular}}
\caption{Component ablations. Each row replaces one part of the default, $C$ by its uniform mean, $\rho$ by equal shares, and the raw update by a drop-and-rescale sparse update. Domain-level values are in the supplementary material.}
\label{tab:ablation}
\end{table}

Sec.~\ref{sec:exp_deficit} showed the allocation behaves as designed. This section asks whether it has to be built this way. Table~\ref{tab:ablation} removes each factor of Eq.~\ref{eq:budget} in turn, which tests whether \textbf{C1} and \textbf{C2} are each necessary. Removing a factor also removes the direction it carries, so the controls that follow hold the opened support fixed and change only direction, one group for each of \textbf{C1}, \textbf{C2}, and \textbf{C3}. Two observations follow. 1)~\emph{Neither factor can be removed.} Replacing $C$ by its uniform mean costs 5.32 to 10.38 points and replacing $\rho$ by equal shares costs 5.09 to 10.92 points, in both cases the same order as the entire paired gain of Sec.~\ref{sec:exp_main}. Collapsing either factor gives up the whole gain, and on Gemma-2-2B-it it costs more than the gain. All six ablated checkpoints land within 2.4 points of the anchor and five of them below it, so each signal carries something the other does not. 2)~\emph{The update parameterization is not what matters.} The sparse-rescaled variant is inconsistent across pools ($-1.92$, $+0.52$, and $-12.18$), whereas uniform capacity and equal domain share are negative in all three. How much magnitude is written into a layer is therefore second-order next to which coordinates are written.

\begin{table*}[t]
\centering
{\small
\setlength{\tabcolsep}{5pt}
\begin{tabular}{llcccccccc}
\hline
Pool & Row & Linear & TA & TIES & DARE & DARE-TIES & Fisher & RegMean & Mean \\
\hline
\multirow{3}{*}{Llama-3.2-3B} & Anchor       & 36.30 & 39.21 & 40.17 & 40.42 & 37.18 & 36.30 & 35.31 & 37.84 \\
             & $+$\method{} & 41.78 & 45.94 & 44.54 & 46.38 & 42.68 & 41.89 & 41.01 & 43.46 \\
             & Gain         & $+5.48$ & $+6.73$ & $+4.37$ & $+5.96$ & $+5.50$ & $+5.59$ & $+5.70$ & $\mathbf{+5.62}$ \\
\hline
\multirow{3}{*}{Llama-3.1-8B-Instruct} & Anchor       & 60.48 & 48.37 & 60.45 & 55.44 & 60.37 & 63.30 & 60.61 & 58.43 \\
                      & $+$\method{} & 65.51 & 57.70 & 64.91 & 62.08 & 65.60 & 67.30 & 65.79 & 64.13 \\
                      & Gain         & $+5.03$ & $+9.33$ & $+4.46$ & $+6.64$ & $+5.23$ & $+4.00$ & $+5.18$ & $\mathbf{+5.70}$ \\
\hline
\multirow{3}{*}{Gemma-2-2B-it} & Anchor       & 48.40 & 35.94 & 46.33 & 43.02 & 47.27 & 48.75 & 48.16 & 45.41 \\
              & $+$\method{} & 56.98 & 45.82 & 55.95 & 52.83 & 55.94 & 56.03 & 56.61 & 54.31 \\
              & Gain         & $+8.58$ & $+9.88$ & $+9.62$ & $+9.81$ & $+8.67$ & $+7.28$ & $+8.45$ & $\mathbf{+8.90}$ \\
\hline
\end{tabular}}
\caption{Macro-5 for each anchor and for \method{} applied on top of it. Every column is a paired comparison under a fixed pool and anchor. All 21 pairs improve. Values are percentages, and domain-level values are in the supplementary material.}
\label{tab:main_gain}
\end{table*}

\paragraph{Directional controls.}
We construct five controls in three groups that preserve the structure of \method{} and change only the signal. 1)~\emph{Capacity direction.} We reverse the slope of Eq.~\ref{eq:capacity}, and separately shuffle the layer profile $C$ across layers. 2)~\emph{Share direction.} We reverse the deficit ordering of Eq.~\ref{eq:share}, and separately sample $\rho$ uniformly from the simplex. 3)~\emph{Coordinate selection.} We write the same number of coordinates per layer and domain with the same $\lambda$, changing only which coordinates are selected. 

\begin{figure}[t]
\centering
\includegraphics[width=\linewidth]{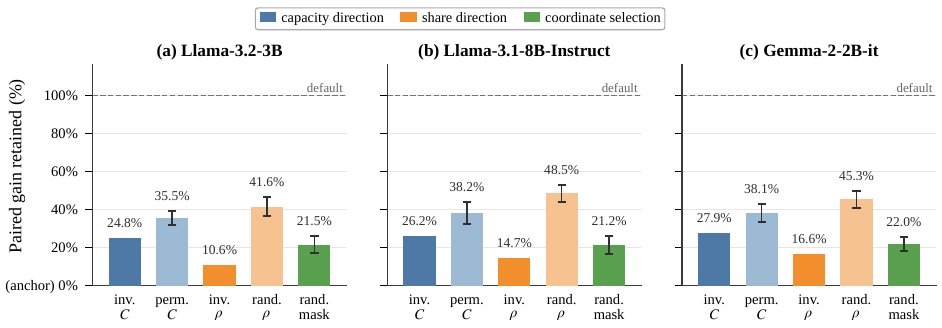}
\caption{Directional controls. Panels (a), (b), and (c) use the DARE, Linear, and Linear anchors. Bars and labels give the percentage of that setting's paired gain a control retains, so $100\%$ is the default \method{} setting and $0\%$ is the anchor. Error bars are the standard deviation over three seeds.}
\label{fig:directional}
\end{figure}

Fig.~\ref{fig:directional} reports what each control retains. Three observations follow. 1)~\emph{No control keeps the gain.} The best of the five retains 42\%, 49\%, and 45\% across the three settings, and the five rank identically in all three. Every control opens the same support as the default, so the shortfall measures the direction alone. 2)~\emph{Inverting a signal costs more than discarding it.} Reversing the capacity slope retains 25\% to 28\% against 35\% to 38\% for a permuted profile, and reversing the deficit ordering retains 11\% to 17\% against 42\% to 49\% for shares drawn from the simplex. The gap between inverting and randomizing is 28 to 34 points for $\rho$ and 10 to 12 points for $C$, so both factors carry a direction and $\rho$ carries the sharper one. This is what the uniform-capacity ablation above could not separate. 3)~\emph{Which coordinates are written matters at a fixed budget.} Equal-budget random masks retain 22\%, 21\%, and 22\%, the tightest spread of the five controls. They write the same number of coordinates per layer and domain at the same $\lambda$ as the default and still give up four fifths of the gain. Which coordinates those are is what the claiming rule of Sec.~\ref{sec:sequential} decides.

\subsection{Paired Comparison with the Anchors}
\label{sec:exp_main}

Table~\ref{tab:main_gain} reports the paired comparison. Four observations follow. 1)~\emph{The gain is universal.} All 21 anchor-pool pairs improve, by $+6.74$ points on average and $15.0\%$ relative, with per-pool relative improvements of $14.9\%$, $10.1\%$, and $19.9\%$. The smallest gain is $+4.00$ and the largest $+9.88$. A Wilcoxon signed-rank test over the 21 pairs gives $p=9.5\times10^{-7}$ (two-sided). 2)~\emph{The gain is stable under reseeding.} For the two stochastic anchors, all 18 anchor-pool-seed pairs improve and the paired gain spans at most $0.12$ points across seeds, well below the smallest gain in the table. 3)~\emph{Anchor strength does not predict the gain.} The largest single gains fall on the weakest anchor, TA, at $+9.33$ on Llama-3.1-8B-Instruct and $+9.88$ on Gemma-2-2B-it, and the relation stops there. On Llama-3.2-3B the stronger and weaker halves of the anchors gain almost identically ($+5.64$ versus $+5.59$), and the smallest gain in each pool falls on TIES or Fisher, neither of which is that pool's weakest anchor. How much room \method{} finds follows from what the anchor failed to retain, not from its aggregate score. 4)~\emph{One pool separates from the other two.} Gemma-2-2B-it improves by $+8.90$ points on average against $+5.62$ and $+5.70$ for the two Llama pools, and it is also where the residual gap falls furthest, $-8.07$ against $-5.60$ and $-5.42$ (Table~\ref{tab:residual_gap}). All three pools start from nearly the same anchor deficit, 12.52, 11.88, and 12.00 points, so what separates them is how much of that deficit one bounded round can reach, not how large it is.

\subsection{Comparison with Baselines}
\label{sec:exp_baselines}

Table~\ref{tab:main_gain} shows the repair improves every anchor it is applied to. Table~\ref{tab:baselines} asks whether a plain dense merge, once repaired, already exceeds methods that carry out both roles themselves. Three observations follow. 1)~\emph{A repaired plain merge exceeds every baseline.} \method{} on a plain dense merge beats the strongest baseline by $+2.46$, $+1.18$, and $+5.74$ points. The same holds on the residual expert gap, where the lowest gap reached by a \method{} checkpoint is 3.94, 3.26, and 1.79 points against 6.64, 4.08, and 6.17 for the best baseline in each pool. Ranking the six methods separately within each pool and domain, \method{} attains the best average rank, 1.67, and takes first place in 11 of the 15 pool-domain cells. 2)~\emph{Search is not a substitute for measuring the deficit.} MERGE$^3$ evolves merge configurations against an item-response-theory proxy and is the strongest baseline on Llama-3.2-3B, yet \method{} applied to DARE exceeds it by $2.46$ points. On Gemma-2-2B-it MERGE$^3$ is the weakest of the five. A search supervises only what its proxy measures (Sec.~\ref{sec:intro}). 3)~\emph{The anchor still matters.} How many \method{} checkpoints clear the strongest baseline varies sharply by pool, 3 of 7 on Llama-3.2-3B, 1 of 7 on Llama-3.1-8B-Instruct, and 6 of 7 on Gemma-2-2B-it. Across all checkpoint-baseline pairs \method{} wins 74 of 105. On Llama-3.1-8B-Instruct the anchors are already strong and CABS trails the best \method{} checkpoint by $1.18$, so the margin there depends on which anchor is chosen.

\begin{table}[t]
\centering
{\small
\setlength{\tabcolsep}{4pt}
\begin{tabular}{lccc}
\hline
Method & \shortstack{Llama\\3.2-3B} & \shortstack{Llama\\3.1-8B-Inst} & \shortstack{Gemma\\2-2B-it} \\
\hline
CABS       & 43.25 & 66.12 & 51.24 \\
CAT        & 41.52 & 63.62 & 49.52 \\
LiNeS      & 41.18 & 62.84 & 49.31 \\
LOT        & 40.90 & 63.18 & 49.02 \\
MERGE$^3$  & 43.92 & 65.58 & 46.20 \\
\hline
Best $+$\method{} & \textbf{46.38} & \textbf{67.30} & \textbf{56.98} \\
Margin            & $+2.46$ & $+1.18$ & $+5.74$ \\
\hline
\end{tabular}}
\caption{Macro-5 of the five baselines against the best checkpoint \method{} produces in each pool.}
\label{tab:baselines}
\end{table}

We attribute the consistent gains of \method{} to three factors. 1)~The allocation follows measured domain deficits (Sec.~\ref{sec:exp_deficit}), so capacity is spent where the anchor falls short (Spearman $+0.76$) instead of spread uniformly. 2)~Conflict-derived layer capacity and deficit-derived domain shares each account for 5 to 11 points on their own (Sec.~\ref{sec:exp_decomposition}), so the outer product supplies structure that neither signal carries alone. 3)~None of the baselines routes on measured per-domain performance. CABS and LiNeS decide from the task vectors and from depth, CAT Merging and LOT Merging from unlabeled activations, and MERGE$^3$ collapses task performance into a single fitness for recipe search. \method{} reads that performance per domain and spends capacity where it is lowest.

\section{Conclusion}
We introduced \method{}, a training-free repair for dense expert merging that allocates bounded support using layer conflict and domain deficits. 
Across 7 anchors and 3 model pools, \method{} improves all 21 paired settings by 15.0\% relative, reduces the average residual expert gap from 12.13 to 5.77 points, and outperforms five baselines. Ablations confirm that both conflict-derived capacity and deficit-driven allocation are essential.

\bibliography{aaai2027}

\input{SupplementaryMaterial.tex}

\end{document}

%% file: SupplementaryMaterial.tex
\clearpage

\section*{Appendix}

\vspace{1em}
\hrule
\vspace{1em}

\appendix
\setcounter{secnumdepth}{2}

\newtheorem{theorem}{Theorem}
\newtheorem{assumption}{Assumption}

\section{What This Supplement Contains}
\label{app:overview}

\method{} is a training-free repair operator. It takes a dense merge of domain experts as its anchor,
measures where that anchor gave capability away, and writes a bounded update back into the coordinates
it can afford to spend. No part of the layer-domain allocation is searched. Every quantity it needs is read
once, from the expert pool, from the task vectors, and from one score per domain.

The main paper states three challenges that allocating capacity has to resolve without training.
\textbf{C1} is deciding where and how many coordinates a layer may give away. \textbf{C2} is deciding
which domains receive them. \textbf{C3} is filling the resulting budget table, one entry per layer and
domain, without searching over merge recipes. The construction answers them with a capacity profile
$C$ read from cross-expert conflict, a share vector $\rho$ read from measured domain deficits, and a
sequential occupancy procedure over the rank-one budget $B=C\rho^{\top}$.

This supplement follows the same three questions. Table~\ref{tab:appmap} maps them onto the sections
that answer them. The optimality statement for the occupancy procedure and its proof sit with
\textbf{C3}, in App.~\ref{app:occupancy}.

\section{Setup}
\label{app:setup}
Everything later in this supplement assumes what is recorded here, namely how the signals are read,
how the benchmarks are split, and what the three constants are set to.

\subsection{Implementation Protocol}
\label{app:implementation}
All merging and evaluation runs use one NVIDIA H100 80GB HBM3 GPU. Every run executes inside a single container image, so the environment is identical across pools, anchors, and baselines. The software stack uses Ubuntu~22.04 with CUDA~12.9, Python~3.10 under conda, PyTorch, and vLLM~0.11.0.

\paragraph{Probe prompts and representation extraction.}
Of the three views in Eq.~\ref{eq:views}, only the representation view $r_i$ reads data. The other
two, $p_i$ and $s_i$, are computed from the task vectors alone. The probe set holds 100 prompts, 20 drawn under a
fixed seed from each of the five MergeBench validation sets and rendered with that domain's own
instruction template. It is built once and reused for every expert, every layer, every anchor, and
every pool, so $r_i$ compares experts on identical inputs. These prompts come from the validation
data the experts were tuned against, not from the eleven benchmarks of
Table~\ref{tab:split_membership}, so the conflict signal never reads an evaluation item. This is a
stronger separation than the calibration split gives $\rho$, which is drawn from the benchmarks
themselves.

\begin{table}[t]
\centering{\small
\setlength{\tabcolsep}{4pt}
\begin{tabular}{p{0.70\linewidth}l}
\hline
Question & Section \\
\hline
What every later section assumes & App.~\ref{app:setup} \\
\textbf{C1} Where and how many coordinates a layer may give away & App.~\ref{app:capacity} \\
\textbf{C2} Which domains receive them & App.~\ref{app:shares} \\
\textbf{C3} How the layer-domain budget is filled without search & App.~\ref{app:occupancy} \\
Whether each of the three is needed & App.~\ref{app:controls} \\
What the repair costs and how far it moves under reseeding & App.~\ref{app:cost} \\
The values behind every Macro-5 number, by domain and by submetric & App.~\ref{app:shares}, App.~\ref{app:results} \\
What the design cannot do & App.~\ref{app:limitations} \\
What each checklist answer rests on & App.~\ref{app:release} \\
\hline
\end{tabular}}
\caption{How this supplement maps onto the three challenges the main paper states.}
\label{tab:appmap}
\end{table}

Each expert is loaded and run once in inference mode with caching disabled, over
prompts tokenized with padding and truncation to 256 tokens. For every transformer block we pool that
block's output hidden states across the non-padding positions with an attention-mask-weighted mean,
giving one $100\times d$ matrix per expert per block. The embedding output is excluded, so layer $i$
in Eq.~\ref{eq:views} is the output of the $i$-th block and no block is subsampled.
$\mathrm{CKA}$ is the biased centered linear variant computed through the $100\times 100$ prompt Gram
matrices and clipped to $[0,1]$, and $r_i$ averages $1-\mathrm{CKA}$ over all ten unordered expert
pairs without weighting. No sampling enters this pass, so $C$ is deterministic given the expert
pool and the probe set.
Because Eq.~\ref{eq:views} reads only the experts and the reference, $C$ is also independent of the
anchor. It is computed once per pool and reused across all seven, which is why the forward-pass count
in Table~\ref{tab:efficiency} does not grow with the number of anchors considered.

Macro-5 is the unweighted mean of the math, instruction, coding, multilingual, and safety domain aggregates. The eleven benchmark variants and their reported submetrics are detailed in App.~\ref{app:submetrics}. HumanEval+ and MBPP+ use the EvalPlus versions, the multilingual benchmarks use the Okapi \texttt{fr}, \texttt{es}, \texttt{de}, and \texttt{ru} versions, HarmBench uses \texttt{text-test}, and DoAnythingNow uses the evaluator JSONL.

Each benchmark is partitioned once into a calibration split and an evaluation split in a $20/80$ ratio, drawn in shuffled order under a fixed seed. The reference $Q_d(\theta_d)$ that Eq.~\ref{eq:gap} measures the
deficit against is the highest score any expert in the pool reaches on domain $d$, which is the
domain's own expert on twelve of the fifteen pool-domain pairs. The same reference is used for the
allocation and for the residual expert gap the main paper reports. Deficits are estimated only on the calibration split, and every score we report comes from the disjoint evaluation split, so no item used to measure a deficit contributes to the number that measures the repair. Each deficit is read on 100 items per domain rather than on the whole
calibration split. The 100 are drawn from that domain's calibration items alone, and where a domain
aggregates more than one benchmark they are divided between those benchmarks in proportion to the
benchmarks' calibration sizes. Five domains therefore account for the 500 labeled samples
Table~\ref{tab:efficiency} counts, and Eq.~\ref{eq:gap} evaluates them once per anchor. The partition is materialised once and reused by every expert, anchor, baseline, and merged checkpoint, which means all methods are compared on identical items rather than on independently drawn samples of the same benchmark.

GSM8K, IFEval, HumanEval+, MBPP+, and the four safety benchmarks are partitioned independently. The three multilingual benchmarks are instead treated as single populations. Their \texttt{fr}, \texttt{es}, \texttt{de}, and \texttt{ru} subsets are concatenated in that fixed order and the benchmark is partitioned once, which holds the language composition of calibration and evaluation identical. Table~\ref{tab:split_membership} lists the resulting sizes, and the counts are the records actually entering the partition, not nominal dataset sizes. Generation-based evaluation uses a fixed engine seed.

\begin{table*}[!t]
\centering{\footnotesize
\setlength{\tabcolsep}{5pt}
\begin{tabular}{lrrr}
\hline
Benchmark variant used & Rows entering split & Calibration & Evaluation \\
\hline
GSM8K & 1,319 & 263 & 1,056 \\
IFEval & 541 & 108 & 433 \\
HumanEval+ (EvalPlus) & 164 & 32 & 132 \\
MBPP+ (EvalPlus) & 378 & 75 & 303 \\
Okapi mMMLU (\texttt{fr/es/de/ru}) & 52,690 & 10,538 & 42,152 \\
Okapi mARC (\texttt{fr/es/de/ru}) & 4,677 & 935 & 3,742 \\
Okapi mHellaSwag (\texttt{fr/es/de/ru}) & 37,352 & 7,470 & 29,882 \\
WildGuardTest (after evaluator filtering) & 749 & 149 & 600 \\
HarmBench (\texttt{text-test}) & 320 & 64 & 256 \\
XSTest & 450 & 90 & 360 \\
DoAnythingNow (evaluator JSONL) & 300 & 60 & 240 \\
\hline
\textbf{Total} & \textbf{98,940} & \textbf{19,784} & \textbf{79,156} \\
\hline
\end{tabular}}
\caption{Exact benchmark variants, record counts, and split sizes used in our experiments.}
\label{tab:split_membership}
\end{table*}

\subsection{Hyperparameter Configurations}
\label{app:configurations}
\method{} exposes three constants. Table~\ref{tab:hyperparameter_protocol} gives their final values together with the one-at-a-time ranges over which each is varied in the diagnostic sweeps.
\begin{table}[t]
\centering{\small
\setlength{\tabcolsep}{5pt}
\begin{tabular}{lcc}
\hline
Parameter & Final value & Diagnostic values \\
\hline
$C_{\min}$ & 0.10 & $\{0,0.05,0.10,0.15\}$ \\
$C_{\max}$ & 0.45 & $\{0.35,0.45,0.55,0.65\}$ \\
$\lambda$ & 0.60 & $\{0.40,0.60,0.80,1.00\}$ \\
\hline
\end{tabular}}
\caption{Final constants and diagnostic values.}
\label{tab:hyperparameter_protocol}
\end{table}
The final values are selected on the Llama-3.2-3B pool and transferred unchanged to the other two pools and all seven anchors, and no pool- or anchor-specific tuning is performed. The Llama-3.1-8B-Instruct sweep in App.~\ref{app:sensitivity} is therefore a transfer check rather than a selection criterion.

\section{Where the Capacity Comes From (\textbf{C1})}
\label{app:capacity}\label{app:layerwise}\label{app:views}
\textbf{C1} asks where and how many coordinates a layer may give away. The answer is the capacity
$C_i$, read from cross-expert conflict, and the score $C_i$ is computed from is itself the mean of
three readings of the same layer. What follows is that profile, and then a test of whether any one reading
would do on its own.

\begin{figure*}[!t]
\centering
\includegraphics[width=0.95\textwidth]{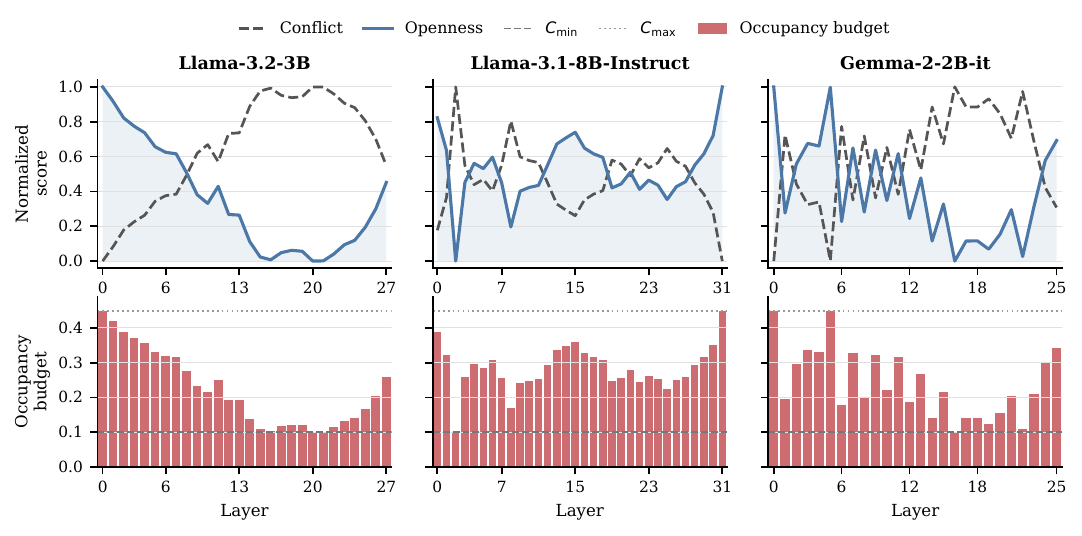}
\caption{Layerwise conflict score $\bar S_i$, the openness it induces, and the resulting occupancy budget $C_i$. The budget decreases with conflict and is clipped to $[C_{\min},C_{\max}]$.}
\label{fig:layerwise}
\end{figure*}

\paragraph{What the profile looks like.}
Fig.~\ref{fig:layerwise} shows the conflict score $\bar S_i$ and the capacity $C_i$ it induces, layer by layer. The profile is not flat, which is what the uniform-capacity ablation of
App.~\ref{app:ablation} removes. It is also not monotone in depth.

\paragraph{The readings do not measure one quantity.}
On Llama-3.2-3B the three readings correlate positively across layers, at $+0.54$, $+0.78$, and
$+0.42$ over the three pairs. On Llama-3.1-8B-Instruct the two weight-space readings still agree at
$+0.60$, while the representation reading runs opposite to both, at $-0.48$ and $-0.38$. Which reading
stands apart is therefore not fixed by the method. A layer whose experts move it in similar directions
with similar signs is not reliably a layer whose function they preserve, and whether those two things
come apart depends on the pool.

\paragraph{No single reading is enough.}
Table~\ref{tab:single_view} builds the capacity profile from one view instead of the mean of three,
holding $C_{\min}$, $C_{\max}$, $\lambda$, $\rho$, and the claiming rule at their defaults. The
uniform-capacity row of App.~\ref{app:ablation} gives the other end of the scale, since it removes the
profile entirely. The best single view recovers about two thirds of what the combined score is worth
and the weakest recovers about one third.

\begin{table}[t]
\centering
{\small
\setlength{\tabcolsep}{5pt}
\begin{tabular}{lccc}
\hline
Capacity built from & Macro-5 & $\Delta$ & recovered \\
\hline
all three views & 46.38 & \phantom{$-$}0.00 & 100.0 \\
sign only & 44.18 & $-2.20$ & 65.9 \\
representation only & 42.66 & $-3.72$ & 42.4 \\
direction only & 42.11 & $-4.27$ & 33.9 \\
uniform capacity & 39.92 & $-6.46$ & \phantom{10}0.0 \\
\hline
\end{tabular}}
\caption{Single-view capacity on Llama-3.2-3B with the DARE anchor. The recovered column is the
percentage of the 6.46 points that the combined profile is worth over uniform capacity.}
\label{tab:single_view}
\end{table}

The three recovered fractions sum to 142 percent, which is consistent with the views carrying
overlapping information and
combining them is not a matter of adding three independent signals. This ablation is run on one pool,
and the correlations above are what caution against reading it as a ranking that transfers. Sign is the
strongest single reading on Llama-3.2-3B, where all three views agree in direction. On
Llama-3.1-8B-Instruct the representation reading is anti-correlated with the other two, and which of them leads there is not something these three pools
separate. Averaging all three needs no answer to that question.

\paragraph{Why the weights are not fitted.}
A weighting that recovered the remaining gap would have to be fitted somewhere, and both places cost
something. Fitting it on the calibration split makes the capacity profile a searched quantity selected
against a proxy, which is the cost \method{} is built to avoid and the ground on which Issue~\#3 of the
main paper separates it from search-based merging. Fitting it per pool gives up the single
configuration that Table~\ref{tab:hyperparameter_protocol} transfers unchanged across three pools and
seven anchors. App.~\ref{app:sensitivity} measures what refusing to tune per pool costs for the three
constants, at most 0.61 Macro-5 points on the one transfer pool we swept against a paired gain of 5.03, and the view
weights are held fixed on the same terms. The measurements above add a third objection, which is that a fit would be made in the
wrong place. The pool we can fit on is Llama-3.2-3B, where all three views correlate positively, so any of them would place capacity
similarly there. Even on that pool Table~\ref{tab:single_view} shows that collapsing onto a single
view costs 2.20 to 4.27 points. The pool where the weighting would matter is
Llama-3.1-8B-Instruct, where the representation view runs opposite to the other two and no single-view
ablation exists to fit against. Equal weights are therefore the aggregation that neither searches for a
weighting nor exports one fitted where the views happened to agree.

\section{Which Domains Receive It (\textbf{C2})}
\label{app:shares}
\textbf{C2} asks which domains receive the coordinates a layer opens. The answer is the share vector
$\rho$, read from measured deficits. Below are the per-domain results those shares produce,
and how far they depend on the split the deficits are read from.

\subsection{Comprehensive Domain-Level Analysis}
\label{app:domainlevel}
\begin{figure*}[!t]
\centering
\includegraphics[width=\textwidth]{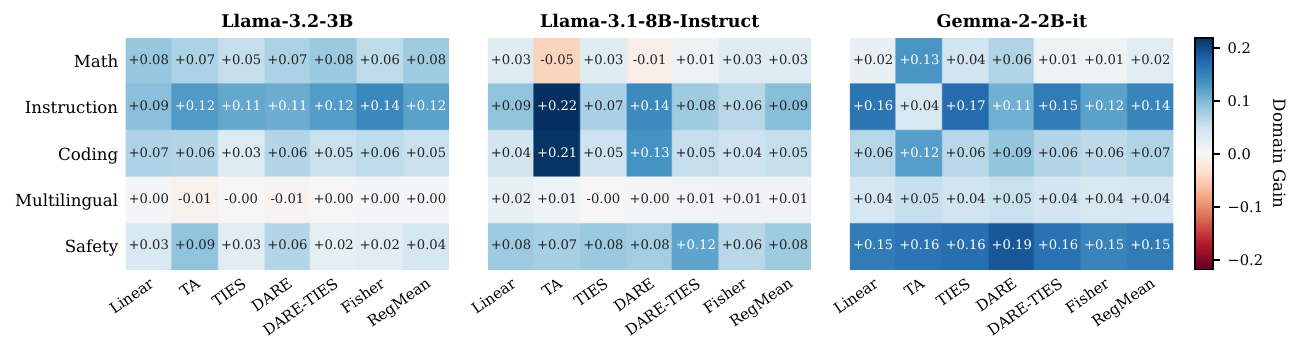}
\caption{Domain gain of \method{} over each dense anchor, for every pool and domain. Each cell is the difference between the $+$\method{} row and the anchor row of Table~\ref{tab:dom_p0} to Table~\ref{tab:dom_p2}. Instruction and safety carry the largest gains. Multilingual is near zero on the two Llama pools and gains 3.5 to 5.0
points on Gemma-2-2B-it.}
\label{fig:domain_gain_matrix}
\end{figure*}

Fig.~\ref{fig:domain_gain_matrix} summarizes all 105 paired domain gains. Tables~\ref{tab:dom_p0}--\ref{tab:dom_p2} then give the exact per-domain values behind every Macro-5 number in the main paper. Each column is one of the seven dense anchors, and the three rows within a domain are the anchor, the checkpoint \method{} builds on it, and their difference.
These are the same 21 paired settings the main paper reports. All 21 improve at the Macro-5 level, but individual domains need not.
Four of the 105 (pool, anchor, domain) triples end further from their expert than the anchor did.

\paragraph{Three of the four are multilingual.}
Two are Llama-3.2-3B settings in which the anchor already matches the multilingual expert, so $\rho$ is
exactly zero, the domain receives no capacity, and its score only drifts by whatever the other domains'
coordinates displace. Over the seven anchors of that pool the drift averages $-0.07$ points, which is
the number the main paper quotes, and on the Task Arithmetic and DARE anchors it carries the score
below the expert and opens gaps of 0.66 and 0.34 points. The third is Llama-3.1-8B-Instruct with the
TIES anchor, where the multilingual deficit is 3.72 points rather than zero and the repair widens it to
3.78. Two further settings score below their own anchor without ending further from the expert, since anchor and repaired checkpoint both stay above it, so the clipped
gap stays at zero.

\paragraph{The fourth is a math regression under one anchor.}
The remaining case is larger and is not a drift. On Llama-3.1-8B-Instruct with the Task Arithmetic anchor, math falls from 78.54 to 73.66, turning a 0.46-point deficit into a 5.34-point one. Math starts only 0.46 points behind on that anchor, so its share of the budget is $0.004$ and it claims almost nothing, which leaves it displaced by whatever the other four domains write. How much
they write is fixed before the anchor is chosen. Eq.~\ref{eq:capacity} reads the capacity profile from
the expert pool alone, so $C$ is the same for all seven anchors of a pool, and Eq.~\ref{eq:share}
normalises $\rho$ to sum to one, so the total support opened is the same for all seven as well. Only the split differs, and a small share is not by itself what produces the failure, since math improves under five of the seven anchors here. Task Arithmetic is also the weakest anchor on this pool, its five deficits summing to 108.9 points against 34.2 to 76.2 for the rest, and seven anchors do not separate that from the share it leaves math.

\paragraph{Where the margin over the baselines comes from.}
Table~\ref{tab:baseline_dom} gives the per-domain values for the five baselines, which the main
paper reports only at the Macro-5 level. Comparing the best checkpoint of each pool against the strongest baseline on each domain
separately, \method{} leads in 11 of the 15 pool-domain cells,
and the four it does not are math twice and multilingual twice. The lead is not spread evenly. On
instruction, safety, and coding it leads on all three pools, by 5.20, 0.47, and 8.09 points on
instruction and by 3.63, 3.24, and 15.10 on safety, the last of these on a cell where no baseline
improves on the anchor by more than 0.23 points. On math and multilingual it leads on one pool
each and trails by 1.64 points at worst. Those five domains rank by aggregate deficit as
instruction, safety, coding, math, multilingual, so the three domains \method{} leads everywhere
are the three the anchors give up most, and the two it splits are the two they give up least.

\begin{table*}[!t]
\centering{\small
\setlength{\tabcolsep}{5pt}
\begin{tabular}{ll rrrrrrr}
\hline
Domain & Row & Linear & TA & TIES & DARE & D-TIES & Fisher & RegMean \\
\hline
Macro-5 & Anchor & 0.3630 & 0.3921 & 0.4017 & 0.4042 & 0.3718 & 0.3630 & 0.3531 \\
 & $+$\method{} & 0.4178 & 0.4594 & 0.4454 & 0.4638 & 0.4268 & 0.4189 & 0.4101 \\
 & Gain & +0.0548 & +0.0673 & +0.0437 & +0.0596 & +0.0550 & +0.0559 & +0.0570 \\
\hline
Math & Anchor & 0.4147 & 0.4799 & 0.5193 & 0.4853 & 0.4488 & 0.4117 & 0.4132 \\
 & $+$\method{} & 0.4967 & 0.5483 & 0.5710 & 0.5536 & 0.5301 & 0.4679 & 0.4907 \\
 & Gain & +0.0820 & +0.0684 & +0.0517 & +0.0683 & +0.0813 & +0.0562 & +0.0775 \\
\hline
Instruction & Anchor & 0.1312 & 0.2165 & 0.1793 & 0.2294 & 0.1165 & 0.1275 & 0.0980 \\
 & $+$\method{} & 0.2226 & 0.3409 & 0.2910 & 0.3372 & 0.2392 & 0.2688 & 0.2152 \\
 & Gain & +0.0914 & +0.1244 & +0.1117 & +0.1078 & +0.1227 & +0.1413 & +0.1172 \\
\hline
Coding & Anchor & 0.3674 & 0.3693 & 0.3921 & 0.3819 & 0.3833 & 0.3754 & 0.3726 \\
 & $+$\method{} & 0.4343 & 0.4337 & 0.4217 & 0.4451 & 0.4318 & 0.4309 & 0.4207 \\
 & Gain & +0.0669 & +0.0644 & +0.0296 & +0.0632 & +0.0485 & +0.0555 & +0.0481 \\
\hline
Multilingual & Anchor & 0.4799 & 0.4778 & 0.4796 & 0.4780 & 0.4795 & 0.4813 & 0.4782 \\
 & $+$\method{} & 0.4817 & 0.4695 & 0.4795 & 0.4727 & 0.4812 & 0.4833 & 0.4816 \\
 & Gain & +0.0018 & -0.0083 & -0.0001 & -0.0053 & +0.0017 & +0.0020 & +0.0034 \\
\hline
Safety & Anchor & 0.4215 & 0.4172 & 0.4383 & 0.4464 & 0.4309 & 0.4190 & 0.4038 \\
 & $+$\method{} & 0.4540 & 0.5044 & 0.4638 & 0.5104 & 0.4518 & 0.4437 & 0.4424 \\
 & Gain & +0.0325 & +0.0872 & +0.0255 & +0.0640 & +0.0209 & +0.0247 & +0.0386 \\
\hline
\end{tabular}}
\caption{Full domain-level paired results on Llama-3.2-3B. Each column is one dense anchor, and the rows within a domain are the anchor, the checkpoint \method{} builds on it, and their difference.}
\label{tab:dom_p0}
\end{table*}

\begin{table*}[!t]
\centering{\small
\setlength{\tabcolsep}{5pt}
\begin{tabular}{ll rrrrrrr}
\hline
Domain & Row & Linear & TA & TIES & DARE & D-TIES & Fisher & RegMean \\
\hline
Macro-5 & Anchor & 0.6048 & 0.4837 & 0.6045 & 0.5544 & 0.6037 & 0.6330 & 0.6061 \\
 & $+$\method{} & 0.6551 & 0.5770 & 0.6491 & 0.6208 & 0.6560 & 0.6730 & 0.6579 \\
 & Gain & +0.0503 & +0.0933 & +0.0446 & +0.0664 & +0.0523 & +0.0400 & +0.0518 \\
\hline
Math & Anchor & 0.7887 & 0.7854 & 0.7985 & 0.8173 & 0.8122 & 0.7720 & 0.7864 \\
 & $+$\method{} & 0.8158 & 0.7366 & 0.8249 & 0.8024 & 0.8234 & 0.8006 & 0.8150 \\
 & Gain & +0.0271 & -0.0488 & +0.0264 & -0.0149 & +0.0112 & +0.0286 & +0.0286 \\
\hline
Instruction & Anchor & 0.3297 & 0.1165 & 0.3630 & 0.1996 & 0.3408 & 0.4110 & 0.3426 \\
 & $+$\method{} & 0.4177 & 0.3350 & 0.4344 & 0.3379 & 0.4177 & 0.4713 & 0.4362 \\
 & Gain & +0.0880 & +0.2185 & +0.0714 & +0.1383 & +0.0769 & +0.0603 & +0.0936 \\
\hline
Coding & Anchor & 0.5811 & 0.2792 & 0.5612 & 0.4445 & 0.5635 & 0.5909 & 0.5747 \\
 & $+$\method{} & 0.6214 & 0.4941 & 0.6064 & 0.5754 & 0.6151 & 0.6353 & 0.6263 \\
 & Gain & +0.0403 & +0.2149 & +0.0452 & +0.1309 & +0.0516 & +0.0444 & +0.0516 \\
\hline
Multilingual & Anchor & 0.5222 & 0.4931 & 0.5208 & 0.5213 & 0.5223 & 0.5395 & 0.5320 \\
 & $+$\method{} & 0.5390 & 0.5022 & 0.5202 & 0.5232 & 0.5284 & 0.5464 & 0.5391 \\
 & Gain & +0.0168 & +0.0091 & -0.0006 & +0.0019 & +0.0061 & +0.0069 & +0.0071 \\
\hline
Safety & Anchor & 0.8025 & 0.7444 & 0.7791 & 0.7894 & 0.7799 & 0.8516 & 0.7950 \\
 & $+$\method{} & 0.8816 & 0.8172 & 0.8596 & 0.8651 & 0.8953 & 0.9115 & 0.8730 \\
 & Gain & +0.0791 & +0.0728 & +0.0805 & +0.0757 & +0.1154 & +0.0599 & +0.0780 \\
\hline
\end{tabular}}
\caption{Full domain-level paired results on Llama-3.1-8B-Instruct.}
\label{tab:dom_p1}
\end{table*}

\begin{table*}[!t]
\centering{\small
\setlength{\tabcolsep}{5pt}
\begin{tabular}{ll rrrrrrr}
\hline
Domain & Row & Linear & TA & TIES & DARE & D-TIES & Fisher & RegMean \\
\hline
Macro-5 & Anchor & 0.4840 & 0.3594 & 0.4633 & 0.4302 & 0.4727 & 0.4875 & 0.4816 \\
 & $+$\method{} & 0.5698 & 0.4582 & 0.5595 & 0.5283 & 0.5594 & 0.5603 & 0.5661 \\
 & Gain & +0.0858 & +0.0988 & +0.0962 & +0.0981 & +0.0867 & +0.0728 & +0.0845 \\
\hline
Math & Anchor & 0.5414 & 0.3465 & 0.5323 & 0.4890 & 0.5482 & 0.5535 & 0.5338 \\
 & $+$\method{} & 0.5584 & 0.4743 & 0.5766 & 0.5539 & 0.5599 & 0.5622 & 0.5569 \\
 & Gain & +0.0170 & +0.1278 & +0.0443 & +0.0649 & +0.0117 & +0.0087 & +0.0231 \\
\hline
Instruction & Anchor & 0.3419 & 0.1996 & 0.2624 & 0.2384 & 0.2994 & 0.3696 & 0.3437 \\
 & $+$\method{} & 0.5009 & 0.2348 & 0.4307 & 0.3438 & 0.4529 & 0.4861 & 0.4880 \\
 & Gain & +0.1590 & +0.0352 & +0.1683 & +0.1054 & +0.1535 & +0.1165 & +0.1443 \\
\hline
Coding & Anchor & 0.3341 & 0.2185 & 0.3345 & 0.2999 & 0.3304 & 0.3338 & 0.3339 \\
 & $+$\method{} & 0.3958 & 0.3381 & 0.3949 & 0.3858 & 0.3936 & 0.3917 & 0.3993 \\
 & Gain & +0.0617 & +0.1196 & +0.0604 & +0.0859 & +0.0632 & +0.0579 & +0.0654 \\
\hline
Multilingual & Anchor & 0.4634 & 0.4140 & 0.4570 & 0.4426 & 0.4605 & 0.4627 & 0.4641 \\
 & $+$\method{} & 0.5012 & 0.4641 & 0.5005 & 0.4917 & 0.5017 & 0.4980 & 0.5013 \\
 & Gain & +0.0378 & +0.0501 & +0.0435 & +0.0491 & +0.0412 & +0.0353 & +0.0372 \\
\hline
Safety & Anchor & 0.7394 & 0.6185 & 0.7306 & 0.6809 & 0.7250 & 0.7178 & 0.7323 \\
 & $+$\method{} & 0.8927 & 0.7800 & 0.8947 & 0.8661 & 0.8887 & 0.8637 & 0.8853 \\
 & Gain & +0.1533 & +0.1615 & +0.1641 & +0.1852 & +0.1637 & +0.1459 & +0.1530 \\
\hline
\end{tabular}}
\caption{Full domain-level paired results on Gemma-2-2B-it.}
\label{tab:dom_p2}
\end{table*}

\begin{table*}[!t]
\centering{\small
\setlength{\tabcolsep}{4pt}
\begin{tabular}{ll rrrrr}
\hline
Pool & Domain & CABS & CAT & LiNeS & LOT & MERGE$^3$ \\
\hline
Llama-3.2-3B & Macro-5 & 0.4325 & 0.4152 & 0.4118 & 0.4090 & 0.4392 \\
 & Math & 0.5064 & 0.4872 & 0.4941 & 0.4936 & 0.5422 \\
 & Instruction & 0.2852 & 0.2524 & 0.2325 & 0.2265 & 0.2654 \\
 & Coding & 0.4130 & 0.3938 & 0.3858 & 0.3842 & 0.4258 \\
 & Multilingual & 0.4838 & 0.4791 & 0.4816 & 0.4804 & 0.4891 \\
 & Safety & 0.4741 & 0.4635 & 0.4650 & 0.4603 & 0.4735 \\
\hline
Llama-3.1-8B-Instruct & Macro-5 & 0.6612 & 0.6362 & 0.6284 & 0.6318 & 0.6558 \\
 & Math & 0.7930 & 0.7661 & 0.7686 & 0.7743 & 0.8167 \\
 & Instruction & 0.4666 & 0.4216 & 0.3946 & 0.3985 & 0.4275 \\
 & Coding & 0.6220 & 0.5950 & 0.5826 & 0.5872 & 0.6226 \\
 & Multilingual & 0.5453 & 0.5382 & 0.5395 & 0.5401 & 0.5469 \\
 & Safety & 0.8791 & 0.8601 & 0.8567 & 0.8589 & 0.8653 \\
\hline
Gemma-2-2B-it & Macro-5 & 0.5124 & 0.4952 & 0.4931 & 0.4902 & 0.4620 \\
 & Math & 0.5712 & 0.5521 & 0.5603 & 0.5597 & 0.5499 \\
 & Instruction & 0.4200 & 0.3874 & 0.3695 & 0.3634 & 0.3089 \\
 & Coding & 0.3616 & 0.3424 & 0.3356 & 0.3340 & 0.3172 \\
 & Multilingual & 0.4675 & 0.4628 & 0.4657 & 0.4645 & 0.4556 \\
 & Safety & 0.7417 & 0.7313 & 0.7344 & 0.7294 & 0.6784 \\
\hline
\end{tabular}}
\caption{Domain-level results for the five baselines. The main paper reports only their Macro-5, and these are the per-domain values behind it. \method{} is not applied on top of these methods because the paired comparison of the main paper is
against the dense-merge families its anchors come from. Extending it to sparse, layer-wise, and
search-derived checkpoints is left to future work.}
\label{tab:baseline_dom}
\end{table*}

\subsection{Sensitivity of the Allocation to the Partition}
\label{app:partition}
All results in this paper use the single partition of App.~\ref{app:implementation}. To check whether the allocation
depends on which partition is drawn, we re-estimated the domain deficits on Llama-3.1-8B-Instruct under
two further $20/80$ draws. The ordering of $\rho$ is unchanged for all seven anchors, so the claiming order of App.~\ref{app:order} would be the same under any of the three draws. The shares themselves move by up to 0.13 against the primary draw, with instruction the least
stable at 0.58, 0.45, and 0.61 on the Linear anchor. We did not merge
or evaluate under these alternative draws.

\section{Filling the Budget (\textbf{C3})}
\label{app:occupancy}
\textbf{C3} asks how a table with one entry per layer and domain gets filled without search. The
answer is sequential occupancy in deficit order. What that procedure optimizes is proved below, followed by what the rank-one
budget does and does not fix, what the claiming order is worth, and what happens as the pool grows.

\subsection{Stagewise Optimality of Sequential Occupancy}
\label{app:theory}
\begin{figure}[t]
\centering
\includegraphics[width=\linewidth]{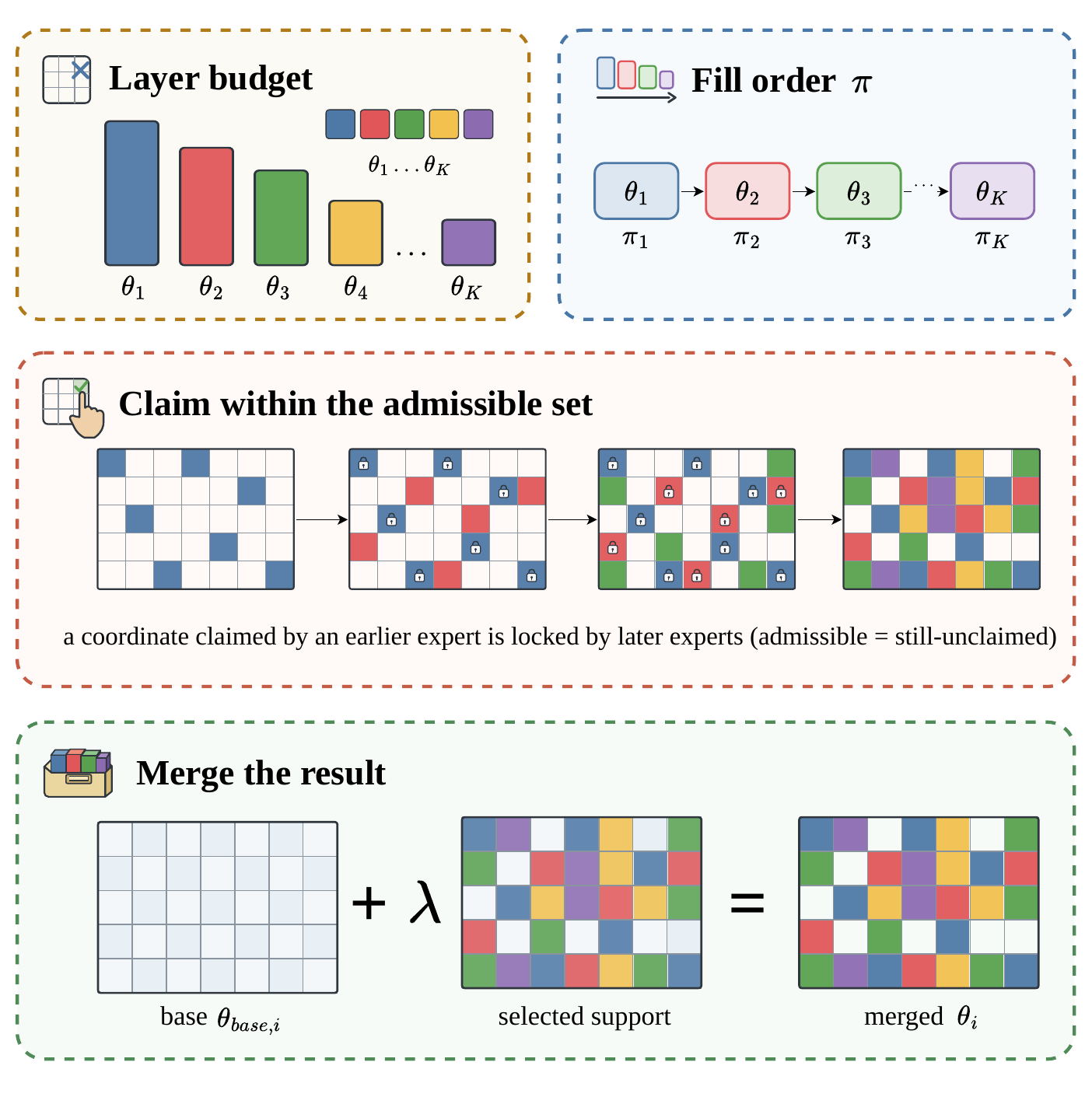}
\caption{Sequential occupancy in one layer. The rounded budget fixes a quota for each domain expert, the deficit order $\pi$ sets precedence, and each expert claims its largest-magnitude coordinates among those still unclaimed. Theorem~\ref{thm:occupancy} establishes quota feasibility and the stagewise optimality of each claim step.}
\label{fig:occupancy}
\end{figure}

\paragraph{Setting.}
Fix a layer, suppress its index, and let $\Omega$ be its finite nonempty coordinate set with $n=|\Omega|$. We consider the nontrivial branch of Alg.~\ref{alg:sigmerge}, where the deficit shares satisfy $\rho_d\geq 0$ and $\sum_d\rho_d=1$. Let $C\in(0,1)$ be the layer capacity target and define the implemented integer quota
\[
k_d=\left\lceil C\rho_d n\right\rceil .
\]
Here $C$ and $C\rho_d$ are pre-rounding fractional targets, rather than strict upper bounds on the realized integer-mask occupancy.
Write $\mathcal D_+=\{d:\rho_d>0\}$, $m_+=|\mathcal D_+|$, and $K=\sum_d k_d$. Let $\pi$ be a fixed total order extending non-increasing $\rho_d$; a deterministic rule resolves equal deficit shares. At step $r$, Alg.~\ref{alg:sigmerge} has available set
\[
R_r=\Omega\setminus\bigcup_{s<r}G_{\pi(s)},
\]
where $G_{\pi(s)}$ is the claimed-coordinate mask from step $s$. For any $M\subseteq\Omega$, define the admitted task-vector mass of domain $d$ by
\[
u_d(M)=\sum_{j\in M}|\Delta_d[j]|.
\]
The sets $G_d$ count claimed coordinates. For the corresponding masked update $V_d=\lambda\mathbf 1_{G_d}\odot\Delta_d$, a selected entry with $\Delta_d[j]=0$ belongs to the claimed mask but not to the algebraic support of $V_d$. Fig.~\ref{fig:occupancy} traces one layer of this procedure. The rounded quotas $k_d$ fix how many coordinates each domain may claim, the deficit order $\pi$ fixes the sequence in which they claim, and each domain in turn takes its largest-magnitude coordinates from whatever remains in $R_r$.

\begin{assumption}
\label{ass:occupancy}
The task-vector entries are finite, and the layer size satisfies
\[
(1-C)n\geq m_+-1.
\]
Ties between equal-magnitude coordinates are resolved by a fixed rule. In the five-domain setting of this paper, where $C\leq C_{\max}=0.45$, $n\geq 8$ is a sufficient condition.
\end{assumption}

\begin{theorem}
\textbf{(Feasible rounding and priority-conditional optimality).}
\label{thm:occupancy}
Under Assumption~\ref{ass:occupancy}, the rounded quotas are feasible: $K\leq n$. The sequential occupancy rule therefore returns pairwise disjoint claimed-coordinate masks satisfying $|G_d|=k_d$ and $|\bigcup_dG_d|=K$. Its realized claimed-mask occupancy obeys
\[
\begin{aligned}
C &\leq \frac{K}{n}<C+\frac{m_+}{n},\\
C\rho_d &\leq\frac{k_d}{n}<C\rho_d+\frac{1}{n}
\quad(d\in\mathcal D_+).
\end{aligned}
\]
The actual algebraic supports satisfy
\[
|\operatorname{supp}(V_d)|\leq k_d,
\qquad
\left|\bigcup_d\operatorname{supp}(V_d)\right|\leq K.
\]
These support bounds hold with equality for every domain and for the union if $\lambda\neq 0$ and $\Delta_d[j]\neq 0$ for every selected pair $(d,j)$ with $j\in G_d$.
Moreover, at every step $r$,
\[
G_{\pi(r)}\in\arg\max_{\substack{M\subseteq R_r\\|M|=k_{\pi(r)}}}u_{\pi(r)}(M).
\]
Thus, conditional on the quotas, precedence order, and earlier claims, no other mask of the same size admits more $\ell_1$ task-vector mass for the current domain. If, at every step with $0<k_{\pi(r)}<|R_r|$, the $k_{\pi(r)}$-th and $(k_{\pi(r)}+1)$-st largest values of $|\Delta_{\pi(r)}[j]|$ on $R_r$ are strictly separated, then $(G_d)_d$ is the unique lexicographic maximizer of
\[
\mathbf u_\pi(M)=
\bigl(u_{\pi(r)}(M_{\pi(r)})\bigr)_{r=1}^{|\mathcal D|}
\]
over all families of pairwise disjoint masks with $|M_d|=k_d$; in particular, it is Pareto efficient for these admitted-mass utilities.
\end{theorem}

\paragraph{Proof.}
For each $d\in\mathcal D_+$, the elementary bound $x\leq\lceil x\rceil<x+1$ gives
\[
C\rho_dn\leq k_d<C\rho_dn+1,
\]
whereas $\rho_d=0$ implies $k_d=0$. Summing over domains and using $\sum_d\rho_d=1$ yields
\[
Cn\leq K<Cn+m_+.
\]
Assumption~\ref{ass:occupancy} gives $Cn+m_+\leq n+1$. Since $K$ is an integer and $K<n+1$, we obtain $K\leq n$. Dividing the preceding inequalities by $n$ proves both occupancy bounds in the theorem.

We next prove feasibility and stagewise optimality along the fixed order. Before step $r$, the earlier masks are pairwise disjoint by construction and contain exactly $\sum_{s<r}k_{\pi(s)}$ coordinates. Hence
\[
|R_r|=n-\sum_{s<r}k_{\pi(s)}
\geq\sum_{s\geq r}k_{\pi(s)}
\geq k_{\pi(r)},
\]
where the first inequality uses $K\leq n$. Thus the requested quota is available at every step. Removing each selected mask from $R_r$ then gives pairwise disjoint masks of the prescribed sizes and $|\bigcup_dG_d|=K$.

By definition, $\operatorname{supp}(V_d)\subseteq G_d$, so $|\operatorname{supp}(V_d)|\leq k_d$ and, using the disjointness of the claimed masks, $|\bigcup_d\operatorname{supp}(V_d)|\leq K$. If $\lambda\neq 0$ and every selected task-vector entry is nonzero, then $\operatorname{supp}(V_d)=G_d$ for every $d$, and all these support bounds are equalities.

Fix a step $r$, and write $d=\pi(r)$ and $k=k_d$. If $k=0$, the empty mask is the only feasible choice. Suppose $k>0$. The rule chooses the $k$ coordinates of $R_r$ having the largest values of $|\Delta_d[j]|$. Let $M\subseteq R_r$ be any other set with $|M|=k$. The sets $G_d\setminus M$ and $M\setminus G_d$ have the same cardinality. Every element of $G_d\setminus M$ has magnitude at least that of every element of $M\setminus G_d$, so these sets can be paired with $|\Delta_d[j]|\geq|\Delta_d[j']|$ in every pair. Summing these inequalities and adding the common contribution from $G_d\cap M$ gives
\[
u_d(G_d)\geq u_d(M).
\]
Because $M$ was arbitrary, $G_d$ attains the claimed stagewise maximum. The argument allows magnitude ties; the fixed tie rule makes the returned mask deterministic but is not needed for the inequality.

The same cardinality calculation shows that every feasible mask considered at a stage can be extended to a complete feasible family: after any masks of the prescribed sizes have been fixed through step $r$, at least $\sum_{s>r}k_{\pi(s)}$ coordinates remain for the later steps.

Finally, suppose the stated strict-separation condition holds. At a step with $k_{\pi(r)}=0$ or $k_{\pi(r)}=|R_r|$, the feasible mask is unique; at every other step, strict separation at the cutoff makes the top-$k_{\pi(r)}$ mask the unique stagewise maximizer. Any lexicographically optimal feasible allocation must maximize its first component, whose unique maximizing mask is $G_{\pi(1)}$. Conditional on that forced choice, its second mask lies in $R_2$ and must equal the unique maximizer $G_{\pi(2)}$. Induction over $r$ forces every mask to equal the greedy mask, proving unique lexicographic optimality. A Pareto improvement would be lexicographically larger at the first domain whose utility improves, which is impossible. \hfill$\diamond$

\paragraph{Proof intuition.}
Ceiling changes each positive fractional quota by less than one coordinate, so its total effect is less than $m_+$ coordinates and the stated layer-size condition preserves feasibility. Once earlier masks are fixed, the current domain faces an ordinary top-$k_d$ problem on $R_r$, where an exchange can only decrease the admitted mass.

\paragraph{Rationale for the deficit order.}
We order the domains by deficit because each later one chooses from a smaller admissible set, so serving the largest deficit first gives the least-recovered domain first access. Theorem~\ref{thm:occupancy} takes the order as given and optimizes within it, so the choice of order is not settled by the theorem. App.~\ref{app:order} measures it directly and finds that reversing the order costs 9.4 to 11.3 percent of the paired gain.

\paragraph{Empirical support.}
The coordinate-selection control uses the same claimed-mask quotas per layer and domain at the same $\lambda$ as the default, changing only which coordinates are selected, and retains 22\%, 21\%, and 22\% of the paired gain across the three settings. Selection therefore accounts for roughly four fifths of the gain at a fixed budget, which is the empirical counterpart of the stagewise choice in Theorem~\ref{thm:occupancy}.

\paragraph{What the rank-one budget constrains.}
$B=C\rho^{\top}$ has rank one, so the budget cannot state that one domain needs capacity at one particular
depth, since the same share $\rho_d$ applies at every layer. The pre-rounding count target
$\widetilde{k}_{i,d}=C_i n_i\rho_d$ factorizes as an outer product, but the implemented quota
$k_{i,d}=\lceil\widetilde{k}_{i,d}\rceil$ need not remain rank one after entrywise ceiling. Instead,
for every $d$ with $\rho_d>0$,
\[
C_i\rho_d\leq \frac{k_{i,d}}{n_i}<C_i\rho_d+\frac{1}{n_i},
\]
while $\rho_d=0$ gives $k_{i,d}=0$. Thus ceiling perturbs each positive cell by less than one
coordinate.

Conditional on this integer quota, the precedence order, and the earlier claims, Theorem~\ref{thm:occupancy}
shows that the rule maximizes the admitted task-vector mass within layer $i$,
\[
u_{i,d}=\sum_{j\in G_{i,d}}\bigl\lvert\Delta_{d,i}[j]\bigr\rvert .
\]
The $\ell_1$ mass of the update actually written is $\lVert V_{d,i}\rVert_1=|\lambda|u_{i,d}$.
Neither quantity is required to factorize or to follow a cross-layer ordering, and both depend on the layer's task-vector
magnitudes and on the coordinates claimed earlier.

The controls of App.~\ref{app:controls} separate the two. The equal-budget random mask holds the entire quota matrix at its default value and changes only which coordinates are
claimed, and it retains 21.5, 21.2, and 22.0 percent of the paired gain. Four fifths of what the
repair achieves therefore comes from the choice of coordinates at the fixed quota matrix induced by
the rank-one pre-rounding target rather than from changing that target.

What remains constrained is the pre-rounding relative share. The rank-one target cannot increase a
domain's share specifically at one depth, although $C_i$, $n_i$, and ceiling can still make its
absolute integer quota vary across layers. A higher-rank budget could express such layer--domain
interactions, and we have not tested one.

\subsection{Claiming Order}
\label{app:order}
Domains claim in descending order of $\rho_d$, so the domain furthest behind picks first. The budget
does not depend on this order, only the sequence in which the domains draw from it. Table~\ref{tab:order}
replaces that sequence while holding $C$, $\rho$, $\lambda$, and the per-domain quotas fixed.

\begin{table*}[t]
\centering
{\small
\setlength{\tabcolsep}{4pt}
\begin{tabular}{lcccc}
\hline
Setting & default & random & reverse & reverse cost \\
\hline
Llama-3.2-3B, DARE & 46.38 & 46.06 & 45.82 & 9.4 \\
Llama-3.1-8B-Instruct, Linear & 65.51 & 65.24 & 65.02 & 9.7 \\
Gemma-2-2B-it, Linear & 56.98 & 56.67 & 56.01 & 11.3 \\
\hline
\end{tabular}}
\caption{Claiming order. Reverse serves the smallest deficit first and is deterministic. Random shuffles
the order and is averaged over seeds $\{0,1,2\}$, with a spread of at most 0.35 Macro-5 points. The last
column is the reverse-order loss as a percentage of that setting's paired gain.}
\label{tab:order}
\end{table*}

Order matters, and the three pools agree on what reversing it costs. Reversing it costs 9.4, 9.7, and 11.3 percent of
the paired gain, and shuffling it costs 3.6 to 5.4 percent. Deficit order is therefore better than an arbitrary order, which is better than the reverse order, and the whole range is about a tenth of what the
repair recovers. This is the quantity Theorem~\ref{thm:occupancy} does not speak to, since it fixes the
order and optimizes within it.

\subsection{Scaling to More Domains}
\label{app:scaling}
Every experiment in this paper uses $|\mathcal D|=5$. What the construction does as $|\mathcal D|$
grows is fixed by the quota rule, which this section reads off.

\paragraph{Disjointness and feasibility do not degrade.}
Alg.~\ref{alg:sigmerge} removes each claimed coordinate from the available set before the next domain
picks, so the masks are disjoint for any number of domains. Overlap is prevented by construction, not
by a margin that shrinks as claimants are added. Feasibility is similarly undemanding.
Assumption~\ref{ass:occupancy} asks for $(1-C)n\geq m_+-1$, so
$n\geq\lceil(|\mathcal D|-1)/(1-C_{\max})\rceil$ suffices: 17 coordinates at $|\mathcal D|=10$ and 35
at $|\mathcal D|=20$, against layer widths in the millions. The rounding overhead in
Theorem~\ref{thm:occupancy} grows as $m_+/n$ and stays negligible on the same comparison.

\paragraph{Available-set depletion is controlled by capacity and rounding.}
A domain claiming late draws from a depleted set, and that handicap might be expected to worsen as
more domains claim ahead of it. For every step $r$, Theorem~\ref{thm:occupancy} gives
\[
\begin{aligned}
\frac{|R_r|}{n}
&\geq 1-\frac{K}{n}\\
&>1-C_i-\frac{m_+}{n}\\
&\geq 1-C_{\max}-\frac{m_+}{n}.
\end{aligned}
\]
Thus capacity and the explicit rounding term control how much of the layer can be unavailable before
a domain chooses. This is a combinatorial availability bound, not a bound on the resulting accuracy
loss or on the value of changing the order. App.~\ref{app:order} measures the latter only at
$|\mathcal D|=5$, so it does not establish how the performance effect scales with the pool size.

\paragraph{What does degrade is the per-domain quota.}
The support budget opened in layer $i$ is exactly $C_i n_i$ for every $\rho$ on the simplex, because
\[
\sum_d B_{i,d}n_i=C_i n_i\sum_d\rho_d=C_i n_i.
\]
This identity describes the continuous pre-rounding budget. Alg.~\ref{alg:sigmerge} realizes it through
the integer claimed-coordinate count
\[
K_i=\sum_d\left\lceil C_i\rho_d n_i\right\rceil,
\qquad
C_i n_i\leq K_i<C_i n_i+m_+,
\]
as bounded by Theorem~\ref{thm:occupancy}, and the algebraic support is no larger than $K_i$.
Dividing the fixed opened budget among $|\mathcal D|$ domains makes the average pre-rounding quota exactly
$C_i n_i/|\mathcal D|$. The average realized integer quota differs from this target by less than
$m_+/|\mathcal D|\leq 1$ coordinate. Thus, at fixed $C_i n_i$, the average pre-rounding amount
available per domain decreases as the pool grows. Holding that target average constant requires the
opened budget to grow in direct proportion to the number of domains.

\paragraph{Raising the ceiling is not free.}
That is the tension five domains cannot settle. Table~\ref{tab:sens_p0} puts $C_{\max}=0.65$ at 2.72 Macro-5 points below the default 0.45 on
Llama-3.2-3B and Table~\ref{tab:sens_p1} puts the same change at 0.55 below it on
Llama-3.1-8B-Instruct, so what lifting the ceiling costs is pool-dependent and can be large. Whether a pool of ten or more is better served by a larger budget, by admitting
only the domains with the largest deficits, or by more than one round of occupancy is a question three
pools of five cannot answer.

\section{Whether All Three Are Needed}
\label{app:controls}
The two budget factors enter as separable terms, so each can be removed outright or kept at full size
and pointed the wrong way while the other is held fixed. The claiming rule that spends the budget
cannot be removed without removing the update, so it is tested by holding the quota matrix at its
default and changing only which coordinates are claimed. Both sets of controls are reported here.

\subsection{Full Component Ablation Values}
\label{app:ablation}
\begin{figure*}[!t]
\centering
\includegraphics[width=\textwidth]{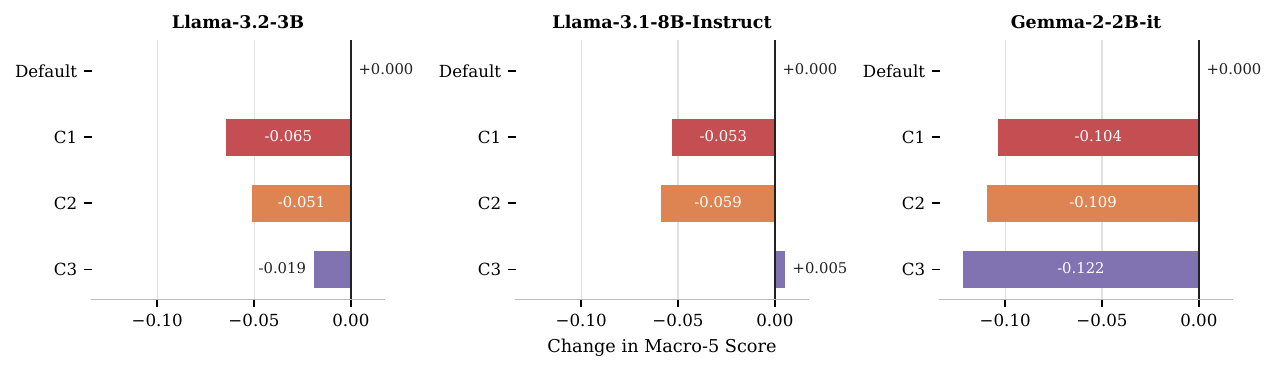}
\caption{Macro-5 cost of removing each component, one panel per pool. The uniform capacity variant replaces the conflict-derived layer capacity with a uniform mean, the equal domain share variant replaces the deficit-driven domain shares with equal shares, and the sparse-rescaled update variant replaces the raw update with a drop-and-rescale sparse update. The first two are negative in all three pools and of comparable size, while the third is not consistently harmful across pools, which separates the two measured signals from the choice of update parameterisation.}
\label{fig:ablation_bar}
\end{figure*}

\begin{figure*}[!t]
\centering
\includegraphics[width=\textwidth]{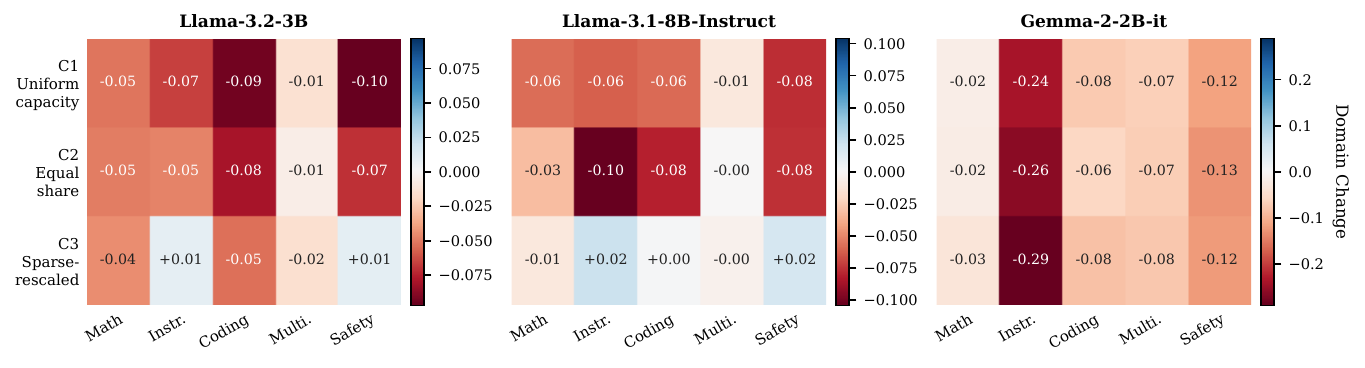}
\caption{Where the cost of each ablation falls, by domain. Averaged over the three pools, removing either measured signal costs most on instruction ($-0.12$ for uniform capacity, $-0.14$ for equal domain share) and safety ($-0.10$, $-0.09$), and least on multilingual ($-0.03$, $-0.03$). Colors are normalized separately within each pool and are not directly comparable across panels. The ordering matches the deficits at the anchor. The domains the anchor gives up most are the ones that suffer when the allocation stops tracking those deficits.}
\label{fig:ablation_domain}
\end{figure*}

Each variant replaces exactly one part of the default and leaves the rest untouched.
Fig.~\ref{fig:ablation_bar} summarizes the Macro-5 effect, Fig.~\ref{fig:ablation_domain} localizes it
by domain, and Table~\ref{tab:ablation_full} gives the exact values.

\paragraph{Both measured signals are load-bearing.}
Replacing $C$ by its uniform mean costs 6.46, 5.32, and 10.38 Macro-5 points across the three pools,
and replacing $\rho$ by equal shares costs 5.09, 5.86, and 10.92. Both are negative on every pool and
of the same order as the paired gain itself, which is 5.96, 5.03, and 8.58. Which of the two costs
more is not fixed. Removing $C$ is worse on Llama-3.2-3B, on all five domains, and removing $\rho$ is
worse on the other two pools, on four of the five domains for Gemma-2-2B-it. Neither factor is the
one that carries the method.

\paragraph{The damage lands where the deficits are.}
Averaged over the three pools, removing either signal costs most on instruction, at 12 and 14 points,
and on safety, at 10 and 9, and least on multilingual, at 3 and 3. That ordering is the ordering of
the deficits at the anchor. The domains the anchor gives up most are the ones that suffer when the
allocation stops tracking what was given up, which is the same mechanism the gains follow in
App.~\ref{app:domainlevel} read in reverse.

\paragraph{The update parameterization is unpredictable rather than small.}
The sparse-rescaled variant gains 0.52 points on Llama-3.1-8B-Instruct and loses 1.92 and 12.18 on the
other two. On Gemma-2-2B-it it is the worst of the three variants, costing more than either signal
ablation, and the loss is concentrated in one domain. Instruction falls from 50.09 to 21.16 there,
which is 13.03 points below the anchor itself, and accounts for 5.79 of the 12.18 Macro-5 points lost.
The same variant raises instruction on both other pools. A choice with this range cannot be
substituted for the two signals, whose ablations agree in sign on every pool and in magnitude to
within 1.4 points of each other.

\begin{table*}[!t]
\centering{\small
\setlength{\tabcolsep}{3pt}
\begin{tabular}{lrrrrrrr}
\hline
Variant & Macro-5 & $\Delta$ & Math & Instr. & Coding & Multi. & Safety \\
\hline
\multicolumn{8}{l}{\emph{Llama-3.2-3B, DARE anchor}} \\
default & 0.4638 & +0.0000 & 0.5536 & 0.3372 & 0.4451 & 0.4727 & 0.5104 \\
uniform capacity & 0.3992 & -0.0646 & 0.5019 & 0.2709 & 0.3515 & 0.4582 & 0.4135 \\
equal domain share & 0.4129 & -0.0509 & 0.5042 & 0.2890 & 0.3653 & 0.4664 & 0.4395 \\
sparse-rescaled update & 0.4446 & -0.0192 & 0.5087 & 0.3457 & 0.3914 & 0.4572 & 0.5200 \\
\hline
\multicolumn{8}{l}{\emph{Llama-3.1-8B-Instruct, Linear anchor}} \\
default & 0.6551 & +0.0000 & 0.8158 & 0.4177 & 0.6214 & 0.5390 & 0.8816 \\
uniform capacity & 0.6019 & -0.0532 & 0.7574 & 0.3559 & 0.5627 & 0.5284 & 0.8052 \\
equal domain share & 0.5965 & -0.0586 & 0.7836 & 0.3137 & 0.5409 & 0.5382 & 0.8059 \\
sparse-rescaled update & 0.6603 & +0.0052 & 0.8052 & 0.4399 & 0.6225 & 0.5347 & 0.8994 \\
\hline
\multicolumn{8}{l}{\emph{Gemma-2-2B-it, Linear anchor}} \\
default & 0.5698 & +0.0000 & 0.5584 & 0.5009 & 0.3958 & 0.5012 & 0.8927 \\
uniform capacity & 0.4660 & -0.1038 & 0.5383 & 0.2634 & 0.3199 & 0.4326 & 0.7760 \\
equal domain share & 0.4606 & -0.1092 & 0.5360 & 0.2393 & 0.3351 & 0.4290 & 0.7638 \\
sparse-rescaled update & 0.4480 & -0.1218 & 0.5239 & 0.2116 & 0.3109 & 0.4234 & 0.7702 \\
\hline
\end{tabular}}
\caption{Full component ablation values. $\Delta$ is Macro-5 against the default row of the same pool.}
\label{tab:ablation_full}
\end{table*}

\subsection{Directional Controls}
\label{app:directional}
The component ablation of App.~\ref{app:ablation} removes a factor outright, which takes away both the
shape of that factor and the direction it points in. The controls here keep the shape and change only
the direction. Table~\ref{tab:directional_full} gives the exact values behind Fig.~\ref{fig:directional}
of the main paper, together with a sixth control that randomizes all three signals at once.

Each control holds the occupancy budget, the sequential claiming rule, and $\lambda$ at their default
values. Inverted capacity reverses the layer ordering of $C$, so the layer that opened most now opens
least. Permuted capacity shuffles $C$ across layers at random. Both keep the multiset of layer
capacities and therefore the total support opened. Inverted share reverses the deficit ordering of
$\rho$ and random share draws $\rho$ uniformly from the simplex, both of which keep $\rho$ on the
simplex. The equal-budget random mask writes the same number of coordinates per layer and domain as the
default and picks them at random. The all-random control applies the permuted, random-share, and
random-mask changes together. The retained fraction is
$(\text{control}-\text{anchor})/(\text{default}-\text{anchor})$, so the default reads $100\%$ and the
anchor reads $0\%$.

\begin{table}[t]
\centering
{\small
\setlength{\tabcolsep}{4pt}
\begin{tabular}{lcccc}
\hline
Control & Macro-5 & sd & $\Delta$ & retained \\
\hline
\multicolumn{5}{l}{\emph{Llama-3.2-3B (DARE). Anchor 40.42, default 46.38}} \\
inverted capacity & 41.90 & 0.00 & $-4.48$ & 24.8 \\
permuted capacity & 42.53 & 0.22 & $-3.85$ & 35.5 \\
inverted share & 41.05 & 0.00 & $-5.33$ & 10.6 \\
random share & 42.90 & 0.30 & $-3.48$ & 41.6 \\
equal-budget random mask & 41.70 & 0.26 & $-4.68$ & 21.5 \\
all three randomized & 40.44 & 0.19 & $-5.94$ & \phantom{$-$}0.3 \\
\hline
\multicolumn{5}{l}{\emph{Llama-3.1-8B-Instruct (Linear). Anchor 60.48, default 65.51}} \\
inverted capacity & 61.80 & 0.00 & $-3.71$ & 26.2 \\
permuted capacity & 62.40 & 0.28 & $-3.11$ & 38.2 \\
inverted share & 61.22 & 0.00 & $-4.29$ & 14.7 \\
random share & 62.92 & 0.22 & $-2.59$ & 48.5 \\
equal-budget random mask & 61.55 & 0.24 & $-3.96$ & 21.2 \\
all three randomized & 60.45 & 0.16 & $-5.06$ & $-0.7$ \\
\hline
\multicolumn{5}{l}{\emph{Gemma-2-2B-it (Linear). Anchor 48.40, default 56.98}} \\
inverted capacity & 50.79 & 0.00 & $-6.19$ & 27.9 \\
permuted capacity & 51.67 & 0.40 & $-5.31$ & 38.1 \\
inverted share & 49.82 & 0.00 & $-7.16$ & 16.6 \\
random share & 52.28 & 0.39 & $-4.70$ & 45.3 \\
equal-budget random mask & 50.28 & 0.32 & $-6.70$ & 22.0 \\
all three randomized & 48.39 & 0.20 & $-8.59$ & $-0.1$ \\
\hline
\end{tabular}}
\caption{Directional controls. Retained fractions are percentages of that setting's paired gain.
Deterministic controls have no seed spread. The randomized single-factor controls use seeds
$\{0,1,2\}$ and the all-random control uses seeds $\{0,\ldots,4\}$.}
\label{tab:directional_full}
\end{table}

Randomizing all three signals returns the checkpoint to its anchor. The three means sit $+0.02$, $-0.03$, and $-0.01$ Macro-5 points from their anchors, 9 of the 15 seeds fall below the anchor, and the 95 percent $t$ interval on each of the three mean
retained fractions lies inside $\pm 4.6$ percentage points. The weakest single-factor control retains 10.6 to 16.6 percent, two to four times that
bound. What the single-factor controls keep therefore comes from the two factors each of them leaves intact, and writing expert coordinates with no signal placing them recovers nothing.

\section{Cost and Reproducibility}
\label{app:cost}
Cost, seed variation, and sensitivity to the three constants follow in that
order.

\subsection{Efficiency}
\label{app:efficiency}
Every run fits on one H100 GPU. Table~\ref{tab:efficiency} separates what each method costs from what
each method has to read.

\paragraph{What each method queries.}
CABS and LiNeS query nothing beyond the task vectors. CAT Merging and LOT Merging take five forward passes over 320 unlabeled exemplars to
read features. That total matches LOT Merging's own recommended budget of 64 per task and is well
above the ten to fifteen CAT Merging asks for. MERGE$^3$ makes 240
evaluation calls against 1500 labeled samples, one call per search iteration, both above the 175
trials and 20 to 100 items its authors use. \method{} sends the probe set through each of the five experts
once for the conflict signal and makes five evaluation calls for the deficits, one per domain. Neither count grows with the number of layers. The main paper's single pass
over the probe prompts is one pass per expert over one shared probe set, which for a five-expert pool
is the five calls Table~\ref{tab:efficiency} records. The conflict signal is read once per pool and
reused across all seven anchors, while Eq.~\ref{eq:gap} measures the deficits on the anchor itself, so those five calls
are paid again for every anchor considered.

\paragraph{Wall-clock.}
The repair adds 0.6 GPU hours to the anchor's 0.4, so a repaired checkpoint costs 1.0 hours. That
places it among the search-free baselines rather than below them, since CABS costs 0.6, LiNeS 0.5,
CAT Merging 1.0, and LOT Merging 1.2. MERGE$^3$ is the outlier at 26.8 hours, 26.8 times the repaired
checkpoint, and the gap is the search loop rather than any per-call cost. Merging itself streams over the task vectors and does not hold the pool in memory, so its memory
footprint is flat in pool size. The probe pass and the deficit calls grow linearly with the number of
domains and the pairwise conflict statistics grow quadratically.

\paragraph{Supervision.}
A merging method can spend labeled data on two separate things, building the merge and fixing its own
free constants, and the table counts only the first. \method{} reads 500 labeled samples from the
calibration split and 100 unlabeled probe prompts to build the merge. On the second it spends one sweep, since $C_{\min}$,
$C_{\max}$, and $\lambda$ are set once on one pool and carried across all three.
CABS and LiNeS build from the task vectors alone and read neither kind of sample, and each then fixes
one scaling coefficient by scoring candidate merges against labeled task accuracy, ten to thirty
candidates in the protocols their papers describe. CAT Merging and LOT Merging read only unlabeled
activations. MERGE$^3$ scores against labeled items at every iteration. Counted this way the repair reads more than four of the five baselines and less
than the fifth. Its three constants were fixed once by a nine-alternative sweep on one pool and one
anchor and are not retuned per pool or per anchor, and that one-time cost sits outside the table on
the same terms as the constants each baseline inherits from its own paper.

\begin{table}[t]
\centering{\footnotesize
\setlength{\tabcolsep}{1.6pt}
\begin{tabular}{lrrrr}
\hline
Method & \shortstack{GPU\\hours} & \shortstack{Calls\\fwd./eval.} & \shortstack{Samples\\lab./unlab.} & \shortstack{Search\\iters.} \\
\hline
Dense anchor  & 0.4  & 0/0   & 0/0    & 0 \\
CABS          & 0.6  & 0/0   & 0/0    & 0 \\
CAT Merging   & 1.0  & 5/0   & 0/320  & 0 \\
LiNeS         & 0.5  & 0/0   & 0/0    & 0 \\
LOT Merging   & 1.2  & 5/0   & 0/320  & 0 \\
MERGE$^3$     & 26.8 & 0/240 & 1500/0 & 240 \\
\hline
$+$\method{}  & 0.6  & 5/5   & 500/100 & 0 \\
\hline
\end{tabular}}
\caption{Cost of building the merge, on one model pool under a single hardware
configuration. Each baseline runs with the constants its own paper reports, so the searches those
papers ran to arrive at them are not counted here. Counts for \method{} are incremental over the
anchor it is applied to.}
\label{tab:efficiency}
\end{table}

\subsection{Randomness and Per-Seed Statistics}
\label{app:perseed}
\method{} is deterministic for a fixed reference, expert pool, anchor, and calibration split. DARE and DARE-TIES sample Bernoulli masks, while MERGE$^3$ performs a stochastic search. Each uses seeds $\{0,1,2\}$. Main-paper tables report seed~0. Table~\ref{tab:perseed} reports the resulting seed-level gains. For the two stochastic \emph{anchors}, DARE and DARE-TIES, the paired gain spans at most $0.0012$ in Macro-5, that is $0.12$ points, well below the smallest gain in the main table. MERGE$^3$ is a standalone baseline rather than a dense anchor, so \method{} is not
applied on top of it, and the lower panel reports its own Macro-5 across seeds instead. Reading the two panels against each other would set a gain beside a score.
Across the six stochastic-anchor settings the repaired checkpoint spans $0.26$
to $0.54$ points over the three seeds, against $0.27$ to $0.46$ for the anchors it is built on, so the repair adds at
most 0.08 points to the anchor's own seed spread. MERGE$^3$ spans $2.88$, $0.83$,
and $3.93$ points on the three pools. Rerunning the search therefore moves the delivered checkpoint
further than resampling an anchor and repairing it does, by roughly a factor of two on
Llama-3.1-8B-Instruct and six to eleven on the other two pools.

The component ablations show the same picture from another angle. Across the three seeds the two deterministic variants are bit-identical, and only the sparse-rescaled variant moves, by $0.4480$, $0.4549$, and $0.4449$ on Gemma-2-2B-it. The spread of that single stochastic row is an order of magnitude smaller than the gap it has to the default.

\begin{table*}[!t]
\centering{\small
\setlength{\tabcolsep}{4pt}
\begin{tabular}{llrrrrrr}
\hline
Pool & Method & seed 0 & seed 1 & seed 2 & mean & std & range \\
\hline
\multicolumn{8}{l}{\emph{Stochastic anchors, paired gain of $+$\method{} over the sampled anchor}} \\
Llama-3.2-3B & DARE & +0.0596 & +0.0599 & +0.0598 & +0.0598 & 0.0002 & 0.0003 \\
Llama-3.2-3B & DARE-TIES & +0.0550 & +0.0544 & +0.0556 & +0.0550 & 0.0006 & 0.0012 \\
Llama-3.1-8B-Instruct & DARE & +0.0664 & +0.0658 & +0.0666 & +0.0663 & 0.0004 & 0.0008 \\
Llama-3.1-8B-Instruct & DARE-TIES & +0.0523 & +0.0519 & +0.0525 & +0.0522 & 0.0003 & 0.0006 \\
Gemma-2-2B-it & DARE & +0.0981 & +0.0978 & +0.0984 & +0.0981 & 0.0003 & 0.0006 \\
Gemma-2-2B-it & DARE-TIES & +0.0867 & +0.0871 & +0.0863 & +0.0867 & 0.0004 & 0.0008 \\
\hline
\multicolumn{8}{l}{\emph{Search-based baseline, Macro-5 across seeds}} \\
Llama-3.2-3B & MERGE$^3$ & 0.4392 & 0.4210 & 0.4498 & 0.4367 & 0.0146 & 0.0288 \\
Llama-3.1-8B-Instruct & MERGE$^3$ & 0.6558 & 0.6641 & 0.6572 & 0.6590 & 0.0044 & 0.0083 \\
Gemma-2-2B-it & MERGE$^3$ & 0.4620 & 0.4412 & 0.4805 & 0.4612 & 0.0197 & 0.0393 \\
\hline
\end{tabular}}
\caption{Seed-level Macro-5 gains. The upper panel reports paired gains for the two stochastic anchors, and the lower panel reports the search-based baseline's Macro-5 across seeds. The two panels therefore report different quantities, a paired gain above and an absolute score below.}
\label{tab:perseed}
\end{table*}

\subsection{Hyperparameter Sensitivity}
\label{app:sensitivity}
\begin{figure*}[t]
\centering
\includegraphics[width=\textwidth]{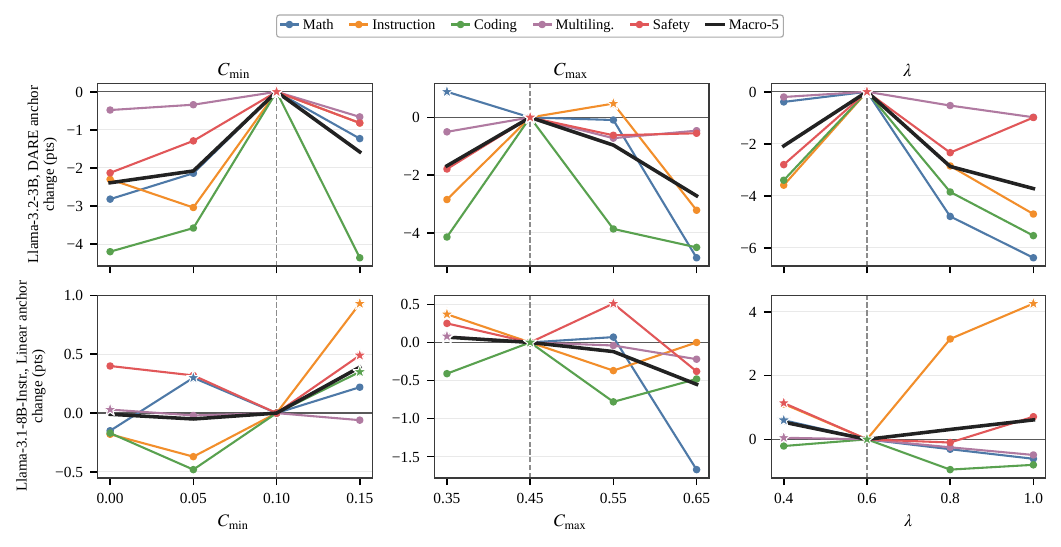}
\caption{Per-domain sensitivity to the three constants. Each panel plots the change from the default
setting, in Macro-5 points, for the five domains and for Macro-5. A star marks each domain's best
value of that constant and the dashed line marks the default. On Llama-3.2-3B every domain peaks at
the default for $C_{\min}$ and $\lambda$, and the stars separate under $C_{\max}$, where math is best
at $0.35$ and instruction at $0.55$. On Llama-3.1-8B-Instruct the Macro-5 line is close to flat while
the domains still move, which is the cancellation reported above.}
\label{fig:sensitivity}
\end{figure*}

Each parameter block varies one constant and holds the other two at their defaults, with the shared
default configuration repeated in bold. Fig.~\ref{fig:sensitivity} gives the overview and
Tables~\ref{tab:sens_p0} and~\ref{tab:sens_p1} give the domain-level values.

\paragraph{What transferring the constants costs.}
The constants are chosen on Llama-3.2-3B and transferred unchanged. On that pool the default is the
best value tested for all three constants, with every one of the nine alternatives scoring lower.
Setting $C_{\min}=0$ costs 2.39 points and $\lambda=1.00$ costs 3.72. On Llama-3.1-8B-Instruct five of
the nine score higher, and the best of them, $\lambda=1.00$, gains 0.61 Macro-5 points. Transferring
rather than tuning therefore costs at most 0.61 points on that pool, against a paired gain of 5.03.

\paragraph{What $\lambda$ trades.}
Macro-5 moves by at most 0.61 points across the $\lambda$ sweep on Llama-3.1-8B-Instruct, and single
domains move much further. Raising $\lambda$ from 0.60 to 1.00 gains instruction 4.26 points and safety
0.71, and costs math, coding, and multilingual 0.61, 0.80, and 0.49. The split follows the deficits at that anchor in sign. The two domains that gain are the two that start furthest behind, at 27.66 and
12.55 points, and the three that lose start at 4.38 or less. A larger $\lambda$ writes more of every
admitted delta, which helps a domain still short of its expert and displaces one already at parity.
A flat aggregate on this pool therefore reflects movements that cancel rather than an absence of them.

\paragraph{Where the domains disagree.}
The five domains do not share an optimum for $C_{\max}$. On Llama-3.2-3B math scores highest at 0.35
and instruction at 0.55, on either side of the default, while coding, multilingual, and safety peak at
the default itself. Since $B=C\rho^{\top}$ gives every domain the same layer profile scaled by its own
share, one global $C_{\max}$ has to serve all five. The default sits at the Macro-5 optimum, and the
compromise costs math 0.89 points and instruction 0.48 against their own best settings. The other two
constants do not create this tension on that pool, where all five domains peak at $C_{\min}=0.10$ and
$\lambda=0.60$.

\begin{table*}[!t]
\centering{\footnotesize
\setlength{\tabcolsep}{3pt}
\begin{tabular}{ll rrrrrrr}
\hline
Group & Changed parameter & Macro-5 & $\Delta$ & Math & Instr. & Coding & Multi. & Safety \\
\hline
floor & $C_{\min}$ = 0.00 & 0.4399 & -0.0239 & 0.5254 & 0.3142 & 0.4031 & 0.4679 & 0.4891 \\
floor & $C_{\min}$ = 0.05 & 0.4430 & -0.0208 & 0.5322 & 0.3068 & 0.4093 & 0.4693 & 0.4975 \\
\textbf{floor} & $\boldsymbol{C_{\min}=0.10}$ & \textbf{0.4638} & \textbf{+0.0000} & \textbf{0.5536} & \textbf{0.3372} & \textbf{0.4451} & \textbf{0.4727} & \textbf{0.5104} \\
floor & $C_{\min}$ = 0.15 & 0.4480 & -0.0158 & 0.5413 & 0.3290 & 0.4015 & 0.4661 & 0.5022 \\
\hline
capacity & $C_{\max}$ = 0.35 & 0.4470 & -0.0168 & 0.5625 & 0.3087 & 0.4036 & 0.4677 & 0.4925 \\
\textbf{capacity} & $\boldsymbol{C_{\max}=0.45}$ & \textbf{0.4638} & \textbf{+0.0000} & \textbf{0.5536} & \textbf{0.3372} & \textbf{0.4451} & \textbf{0.4727} & \textbf{0.5104} \\
capacity & $C_{\max}$ = 0.55 & 0.4542 & -0.0096 & 0.5527 & 0.3420 & 0.4064 & 0.4655 & 0.5042 \\
capacity & $C_{\max}$ = 0.65 & 0.4366 & -0.0272 & 0.5049 & 0.3050 & 0.4000 & 0.4681 & 0.5049 \\
\hline
lambda & $\lambda$ = 0.40 & 0.4430 & -0.0208 & 0.5497 & 0.3013 & 0.4111 & 0.4707 & 0.4824 \\
\textbf{lambda} & $\boldsymbol{\lambda=0.60}$ & \textbf{0.4638} & \textbf{+0.0000} & \textbf{0.5536} & \textbf{0.3372} & \textbf{0.4451} & \textbf{0.4727} & \textbf{0.5104} \\
lambda & $\lambda$ = 0.80 & 0.4351 & -0.0287 & 0.5057 & 0.3087 & 0.4066 & 0.4674 & 0.4870 \\
lambda & $\lambda$ = 1.00 & 0.4266 & -0.0372 & 0.4898 & 0.2902 & 0.3898 & 0.4629 & 0.5005 \\
\hline
\end{tabular}}
\caption{Hyperparameter sensitivity on Llama-3.2-3B with the DARE anchor. Bold rows mark the default value for each parameter, and $\Delta$ is measured against the shared default configuration.}
\label{tab:sens_p0}
\end{table*}

\begin{table*}[!t]
\centering{\footnotesize
\setlength{\tabcolsep}{3pt}
\begin{tabular}{ll rrrrrrr}
\hline
Group & Changed parameter & Macro-5 & $\Delta$ & Math & Instr. & Coding & Multi. & Safety \\
\hline
floor & $C_{\min}$ = 0.00 & 0.6550 & -0.0001 & 0.8143 & 0.4159 & 0.6197 & 0.5393 & 0.8856 \\
floor & $C_{\min}$ = 0.05 & 0.6546 & -0.0005 & 0.8188 & 0.4140 & 0.6166 & 0.5388 & 0.8848 \\
\textbf{floor} & $\boldsymbol{C_{\min}=0.10}$ & \textbf{0.6551} & \textbf{+0.0000} & \textbf{0.8158} & \textbf{0.4177} & \textbf{0.6214} & \textbf{0.5390} & \textbf{0.8816} \\
floor & $C_{\min}$ = 0.15 & 0.6590 & +0.0039 & 0.8180 & 0.4270 & 0.6249 & 0.5384 & 0.8865 \\
\hline
capacity & $C_{\max}$ = 0.35 & 0.6558 & +0.0007 & 0.8165 & 0.4214 & 0.6173 & 0.5398 & 0.8841 \\
\textbf{capacity} & $\boldsymbol{C_{\max}=0.45}$ & \textbf{0.6551} & \textbf{+0.0000} & \textbf{0.8158} & \textbf{0.4177} & \textbf{0.6214} & \textbf{0.5390} & \textbf{0.8816} \\
capacity & $C_{\max}$ = 0.55 & 0.6539 & -0.0012 & 0.8165 & 0.4140 & 0.6136 & 0.5386 & 0.8867 \\
capacity & $C_{\max}$ = 0.65 & 0.6496 & -0.0055 & 0.7991 & 0.4177 & 0.6166 & 0.5368 & 0.8778 \\
\hline
lambda & $\lambda$ = 0.40 & 0.6605 & +0.0054 & 0.8218 & 0.4288 & 0.6193 & 0.5395 & 0.8930 \\
\textbf{lambda} & $\boldsymbol{\lambda=0.60}$ & \textbf{0.6551} & \textbf{+0.0000} & \textbf{0.8158} & \textbf{0.4177} & \textbf{0.6214} & \textbf{0.5390} & \textbf{0.8816} \\
lambda & $\lambda$ = 0.80 & 0.6582 & +0.0031 & 0.8127 & 0.4492 & 0.6119 & 0.5365 & 0.8806 \\
lambda & $\lambda$ = 1.00 & 0.6612 & +0.0061 & 0.8097 & 0.4603 & 0.6134 & 0.5341 & 0.8887 \\
\hline
\end{tabular}}
\caption{Hyperparameter sensitivity on Llama-3.1-8B-Instruct with the Linear anchor. Bold rows mark the default value for each parameter, and $\Delta$ is measured against the shared default configuration.}
\label{tab:sens_p1}
\end{table*}

\section{Complete Results}
\label{app:results}
The main paper reports Macro-5. The values underneath it follow, by anchor, by domain, and
by submetric.

\subsection{Anchor Selection}
\label{app:anchorselect}
\method{} is applied on top of a dense merge, and the seven anchors of a pool differ by up to 15 Macro-5
points before any repair. A practitioner has to pick one. The question here is whether that choice can be made from quantities available
before running \method{} at all.

\paragraph{What we tested.}
For each of the 21 pool-anchor settings we computed nine statistics from the anchor alone, its Macro-5,
the mean, maximum, minimum, and standard deviation of its five residual expert gaps, the number of
domains whose gap is zero, the entropy and Gini coefficient of the normalized deficit vector $\rho$, and
the range of its five domain scores. We correlated each against the paired gain, against the post-repair
Macro-5, and against the post-repair residual gap. Correlations are reported within each pool, since the
three pools differ enough that a pooled correlation can carry the opposite sign from every pool it
averages. 

\paragraph{No statistic predicts the gain on all three pools.}
Every one of the nine reverses sign or collapses to zero on at least one pool. Anchor Macro-5 is the clearest case. It correlates $-0.86$ and $-0.96$ with the gain on two pools and
$+0.11$ on the third, so a rule fitted on one pool would not carry to another, which is the main-paper
observation that anchor strength does not predict how much room \method{} finds. The strongest pooled candidate is the entropy of $\rho$
at $+0.78$ over the 21 settings, but it is $+0.07$ within Llama-3.1-8B-Instruct, where instruction is the largest
share under all seven anchors and the entropy spans only 0.12 nats, so the statistic has little to
separate. Seven anchors per pool is a small sample and the nine
statistics are not independent, since within a fixed pool the mean residual gap is an affine function
of Macro-5. A predictor that needs more anchors, or a statistic we did not compute, would not surface
in a sweep this size.

\paragraph{Anchor strength does predict the final level.}
The range of the anchor's five domain scores correlates $+0.73$ with post-repair Macro-5 over the 21
settings, though it turns negative inside every pool, at $-0.86$, $-0.64$, and $-0.54$. The statistic
that does survive the split is the anchor's own score. Within each pool, anchor Macro-5 and post-repair
Macro-5 correlate at $+0.96$, $+0.89$, and $+0.86$, and
56 of the 63 within-pool anchor pairs keep their order across the repair. Choosing the strongest
available anchor therefore costs little against choosing with hindsight.
Table~\ref{tab:anchor_regret} prices that choice against the three selection rules a practitioner could follow. Taking the strongest dense merge gives up 0.32 Macro-5 points on average against the
anchor that turns out best after the repair, while taking one at random gives up 2.67 to 3.17 and the worst choice gives up
5.37 to 11.16.

\begin{table}[t]
\centering
{\small
\setlength{\tabcolsep}{5pt}
\begin{tabular}{lccc}
\hline
Selection rule & L3.2-3B & L3.1-8B & Gemma \\
\hline
Strongest anchor & 0.00 & 0.00 & 0.95 \\
Random anchor & 2.92 & 3.17 & 2.67 \\
Weakest anchor & 5.37 & 9.60 & 11.16 \\
\hline
\end{tabular}}
\caption{Macro-5 points given up against the best anchor available in that pool, judged after the repair.
Picking the strongest dense merge costs 0.32 points on average.}
\label{tab:anchor_regret}
\end{table}

Part of this is arithmetic, since the post-repair score is the anchor plus the gain and the anchor varies
more than the gain does. The rest is not arithmetic. The gain is strongly negatively correlated with
anchor strength on two of the three pools, so the repair does compress the spread, and the compression
stops short of reordering the anchors. The practical reading is that the repair is close to rank preserving,
so a strong anchor stays strong, while how much it will improve remains unpredictable.

\subsection{Five-Domain Profiles}
\label{app:radar}
\begin{figure*}[!p]
\centering
\includegraphics[width=\textwidth]{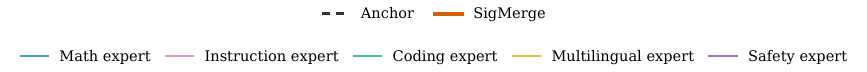}\\[0.7em]
\includegraphics[width=\textwidth]{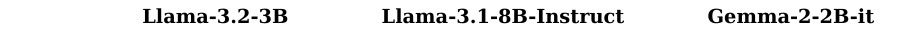}\\[0.7em]
\includegraphics[width=\textwidth]{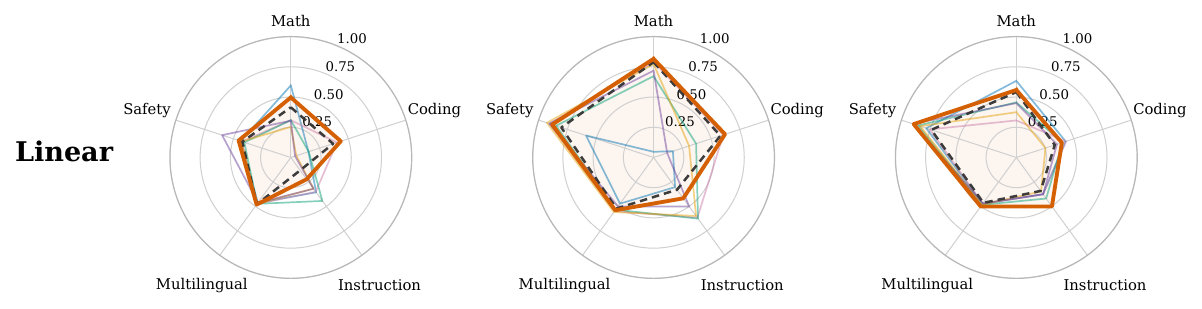}\\
\includegraphics[width=\textwidth]{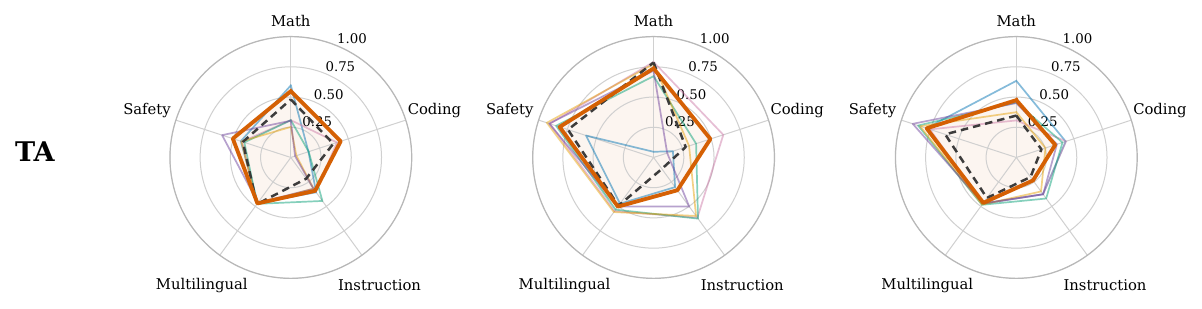}\\
\includegraphics[width=\textwidth]{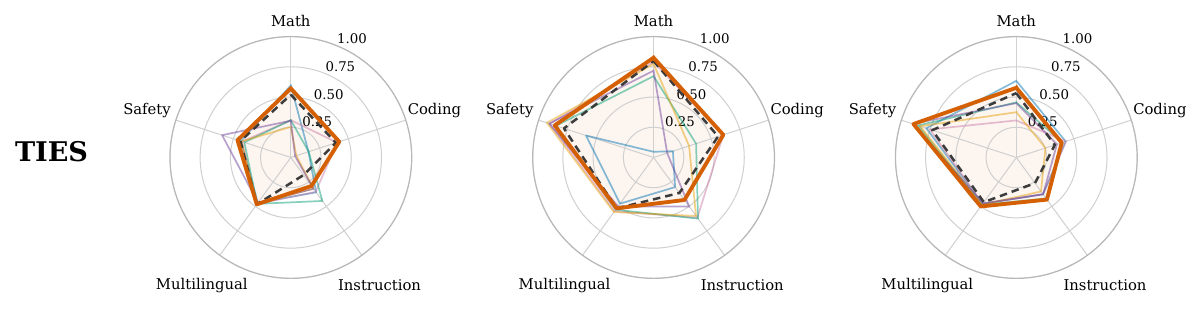}\\
\includegraphics[width=\textwidth]{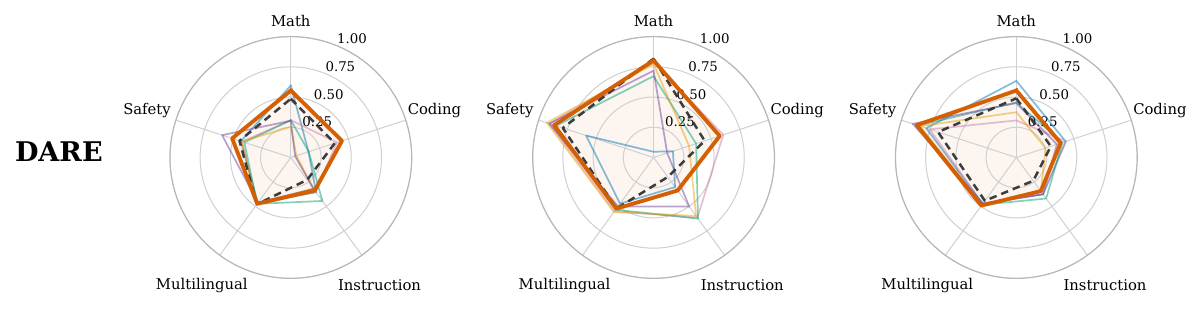}
\caption{Five-domain profiles for the Linear, Task Arithmetic, TIES, and DARE anchors. Columns are the three pools.}
\label{fig:radar1}
\end{figure*}

\begin{figure*}[!p]
\centering
\includegraphics[width=\textwidth]{Figures/style_probe_legend}\\[0.7em]
\includegraphics[width=\textwidth]{Figures/p012_column_titles}\\[0.7em]
\includegraphics[width=\textwidth]{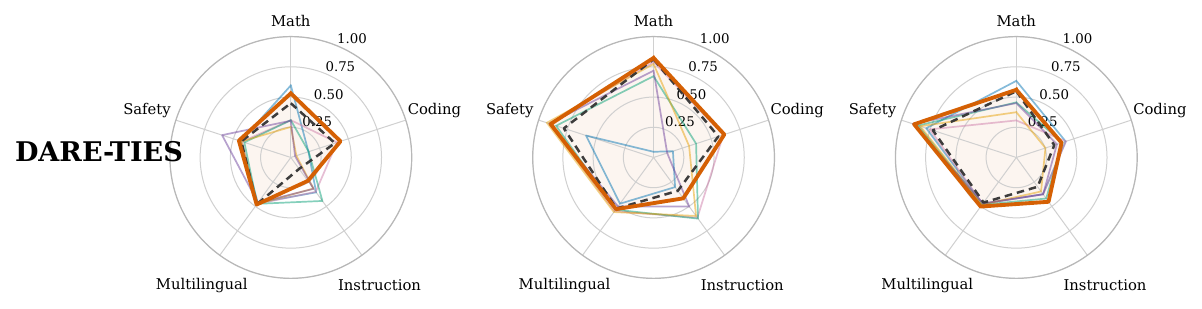}\\
\includegraphics[width=\textwidth]{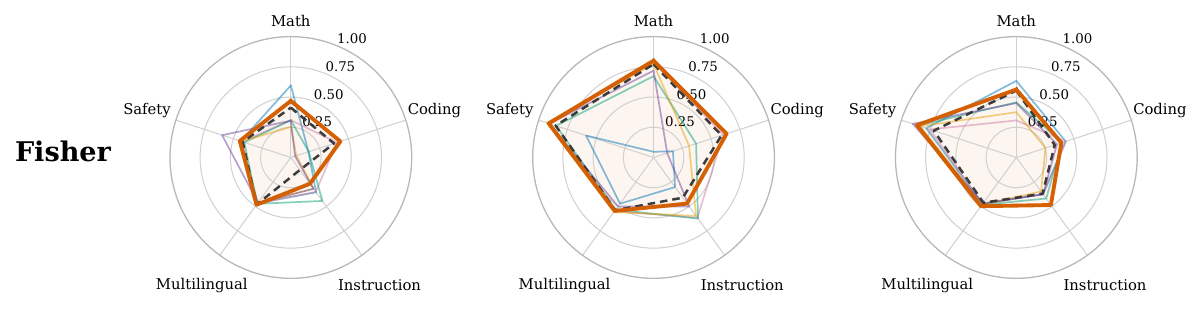}\\
\includegraphics[width=\textwidth]{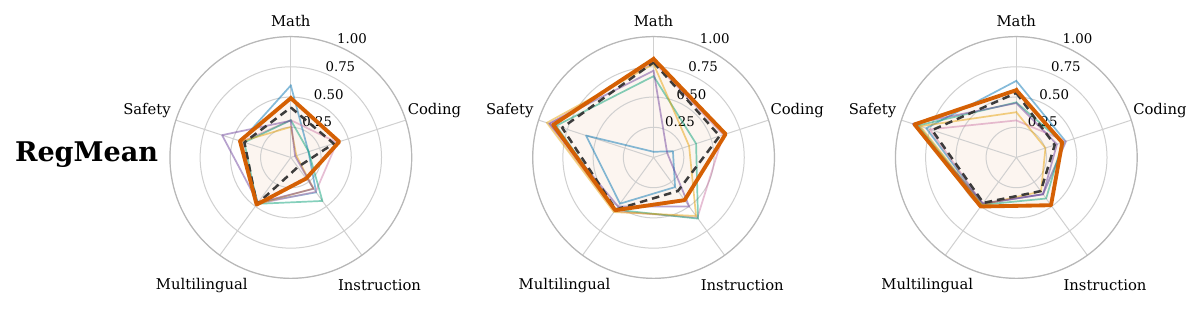}
\caption{Five-domain profiles for the DARE-TIES, Fisher, and RegMean anchors.}
\label{fig:radar2}
\end{figure*}

Fig.~\ref{fig:radar1} and Fig.~\ref{fig:radar2} show the shape of the five-domain profile for every anchor. Each panel draws the five domain experts, the dense anchor, and the checkpoint \method{} builds on that anchor, on the same five axes. Three patterns are visible. The domain-specialized experts are generally lopsided, although the strongest checkpoint on a domain need not be its nominal expert. The anchor pulls those profiles into one shape that is flatter but sits inside the expert envelope on the contested axes. \method{} pushes that shape back outward, and it does so unevenly. The axes that move are the ones the anchor gave up most, matching the deficit-driven allocation geometrically.

\begin{figure*}[!t]
\centering
\includegraphics[width=\textwidth]{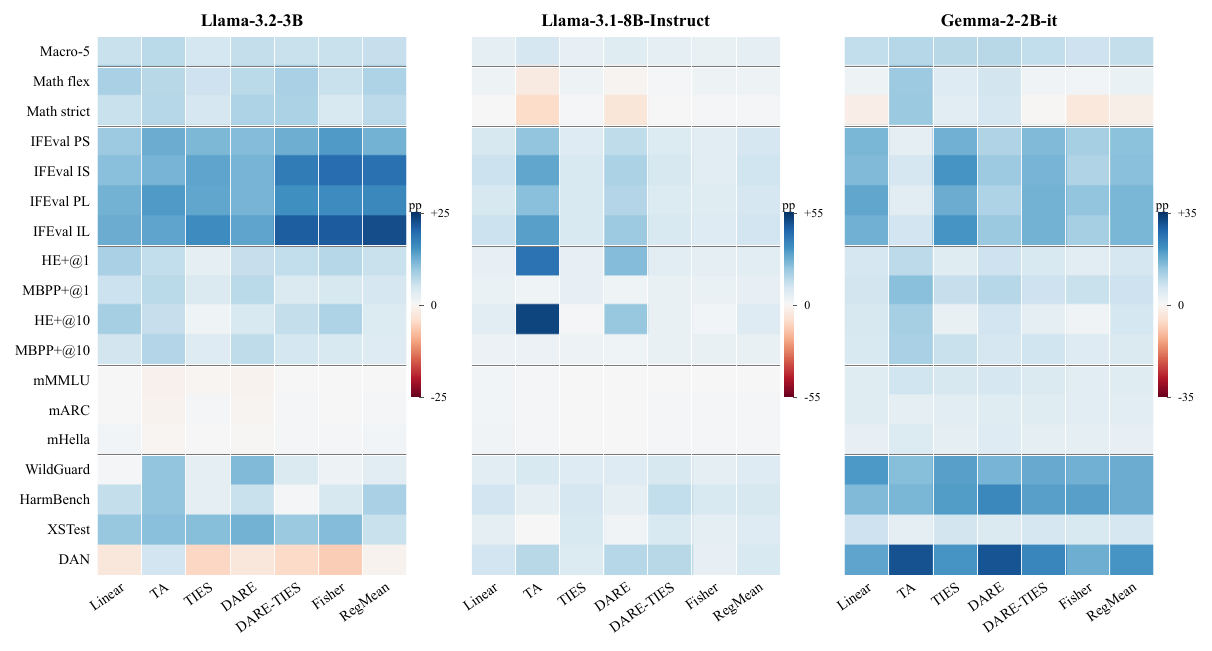}
\caption{Gains on 17 reported metrics from eleven benchmarks, together with Macro-5, one column per anchor and pool. Colors are normalized separately within each pool and are not directly comparable across panels. Improvements are broad rather than concentrated. They appear on both IFEval strictness variants and at pass@1 as well as pass@10, WildGuardTest and HarmBench improve on all 21 settings and XSTest on all but one, where it
is unchanged.}
\label{fig:submetric_heatmap}
\end{figure*}

\subsection{Submetric-Level Results}
\label{app:submetrics}
The Macro-5 aggregate is built from eleven benchmarks. Tables~\ref{tab:sub_math}--\ref{tab:sub_safety} give the exact score after \method{} is applied, with the change against the anchor in parentheses, and Fig.~\ref{fig:submetric_heatmap} summarizes all 17 reported benchmark metrics.

\paragraph{Metric construction.}
Math uses GSM8K flexible extraction as its domain aggregate and reports strict extraction as an additional diagnostic. Instruction uses IFEval prompt-strict accuracy as the domain aggregate and also reports instruction-strict, prompt-loose, and instruction-loose accuracy. Coding averages HumanEval+ and MBPP+ pass@1 for the domain aggregate and reports pass@1 and pass@10 for each benchmark separately. Multilingual averages the Okapi mMMLU, mARC, and mHellaSwag scores after the fixed \texttt{fr/es/de/ru} concatenation described in App.~\ref{app:implementation}. Safety averages WildGuardTest, HarmBench \texttt{text-test}, XSTest, and the DoAnythingNow evaluator JSONL. Macro-5 then gives these five domain aggregates equal weight rather than
weighting them by their number of records or reported diagnostics.

\paragraph{Coverage and deltas.}
Every submetric table contains the same 21 paired settings, seven dense anchors for each of the
three model pools. A parenthesized value is the score after \method{} minus the score of that exact anchor on
the same evaluation membership. It is not a difference against an expert or against a separately selected reference. The exact split populations and shared membership are recorded in Table~\ref{tab:split_membership}.

\paragraph{Where the improvement is uniform.}
Instruction improves on all 21 settings on all four IFEval variants, with mean gains from 11.65 to
15.03 points, and coding improves on all 21 settings on both benchmarks at both sampling budgets.
Pass@1 moves by 8.28 and 5.72 points on HumanEval+ and MBPP+, against 7.82 and
4.95 at pass@10, so the top-ranked completion improves as much as the sampled set does and the coding
gain does not come from drawing more candidates.

\paragraph{Where it is not.}
Two submetrics move less reliably than the domain aggregate they sit under. GSM8K strict extraction
is negative on 7 of the 21 settings against 2 for flexible extraction, and gains 2.04 points on
average against 4.01. DoAnythingNow is negative on 6 of 21 while WildGuardTest and HarmBench are positive on every setting and
XSTest on all but one, where it is unchanged. Multilingual is small throughout, with means between 1.53 and 1.75
points and between 2 and 7 negative settings per submetric, which is the behaviour the near-zero
multilingual deficits of App.~\ref{app:domainlevel} lead to.

\begin{table*}[!t]
\centering{\small
\setlength{\tabcolsep}{3pt}
\begin{tabular}{lcc}
\hline
Anchor & GSM8K flex & GSM8K strict \\
\hline
\multicolumn{3}{l}{\emph{Llama-3.2-3B}} \\
Linear & 0.4967 (+0.0820) & 0.4224 (+0.0562) \\
TA & 0.5483 (+0.0684) & 0.4346 (+0.0714) \\
TIES & 0.5710 (+0.0517) & 0.4732 (+0.0426) \\
DARE & 0.5536 (+0.0683) & 0.4490 (+0.0781) \\
DARE-TIES & 0.5301 (+0.0813) & 0.4566 (+0.0783) \\
Fisher & 0.4679 (+0.0562) & 0.3967 (+0.0381) \\
RegMean & 0.4907 (+0.0775) & 0.4156 (+0.0646) \\
\hline
\multicolumn{3}{l}{\emph{Llama-3.1-8B-Instruct}} \\
Linear & 0.8158 (+0.0271) & 0.7377 (+0.0006) \\
TA & 0.7366 (-0.0488) & 0.6653 (-0.0997) \\
TIES & 0.8249 (+0.0264) & 0.7923 (+0.0044) \\
DARE & 0.8024 (-0.0149) & 0.7084 (-0.0672) \\
DARE-TIES & 0.8234 (+0.0112) & 0.7862 (-0.0002) \\
Fisher & 0.8006 (+0.0286) & 0.7149 (+0.0127) \\
RegMean & 0.8150 (+0.0286) & 0.7377 (+0.0082) \\
\hline
\multicolumn{3}{l}{\emph{Gemma-2-2B-it}} \\
Linear & 0.5584 (+0.0170) & 0.5129 (-0.0224) \\
TA & 0.4743 (+0.1278) & 0.4553 (+0.1308) \\
TIES & 0.5766 (+0.0443) & 0.5524 (+0.0361) \\
DARE & 0.5539 (+0.0649) & 0.5357 (+0.0580) \\
DARE-TIES & 0.5599 (+0.0117) & 0.5319 (-0.0049) \\
Fisher & 0.5622 (+0.0087) & 0.5107 (-0.0367) \\
RegMean & 0.5569 (+0.0231) & 0.5061 (-0.0216) \\
\hline
\end{tabular}}
\caption{Submetric scores on the math domain after applying \method{}, with the change against the anchor in parentheses.}
\label{tab:sub_math}
\end{table*}

\begin{table*}[!t]
\centering{\small
\setlength{\tabcolsep}{4pt}
\begin{tabular}{llcccc}
\hline
Pool & Anchor & Prompt str. & Inst. str. & Prompt loose & Inst. loose \\
\hline
Llama-3.2-3B & Linear & 0.2226 (+0.0914) & 0.3182 (+0.1048) & 0.2817 (+0.1190) & 0.3697 (+0.1239) \\
 & TA & 0.3409 (+0.1244) & 0.4548 (+0.1159) & 0.4204 (+0.1410) & 0.5352 (+0.1328) \\
 & TIES & 0.2910 (+0.1117) & 0.3997 (+0.1323) & 0.3575 (+0.1301) & 0.4692 (+0.1574) \\
 & DARE & 0.3372 (+0.1078) & 0.4608 (+0.1171) & 0.4111 (+0.1170) & 0.5400 (+0.1328) \\
 & DARE-TIES & 0.2392 (+0.1227) & 0.3457 (+0.1742) & 0.2984 (+0.1524) & 0.3973 (+0.2055) \\
 & Fisher & 0.2688 (+0.1413) & 0.3649 (+0.1910) & 0.3224 (+0.1579) & 0.4153 (+0.2079) \\
 & RegMean & 0.2152 (+0.1172) & 0.3158 (+0.1863) & 0.2743 (+0.1615) & 0.3661 (+0.2210) \\
\hline
Llama-3.1-8B-Instruct & Linear & 0.4177 (+0.0880) & 0.5456 (+0.1180) & 0.4640 (+0.0900) & 0.5839 (+0.1179) \\
 & TA & 0.3350 (+0.2185) & 0.4461 (+0.2842) & 0.3627 (+0.2315) & 0.4736 (+0.2997) \\
 & TIES & 0.4344 (+0.0714) & 0.5516 (+0.0820) & 0.5065 (+0.0844) & 0.6163 (+0.0856) \\
 & DARE & 0.3379 (+0.1383) & 0.4529 (+0.1747) & 0.3767 (+0.1604) & 0.4900 (+0.1986) \\
 & DARE-TIES & 0.4177 (+0.0769) & 0.5444 (+0.0868) & 0.4769 (+0.0770) & 0.5971 (+0.0819) \\
 & Fisher & 0.4713 (+0.0603) & 0.5935 (+0.0616) & 0.5213 (+0.0659) & 0.6391 (+0.0712) \\
 & RegMean & 0.4362 (+0.0936) & 0.5647 (+0.1107) & 0.4954 (+0.0936) & 0.6127 (+0.1059) \\
\hline
Gemma-2-2B-it & Linear & 0.5009 (+0.1590) & 0.6079 (+0.1555) & 0.5416 (+0.1812) & 0.6427 (+0.1687) \\
 & TA & 0.2348 (+0.0352) & 0.3441 (+0.0599) & 0.2514 (+0.0388) & 0.3705 (+0.0671) \\
 & TIES & 0.4307 (+0.1683) & 0.5564 (+0.2059) & 0.4547 (+0.1738) & 0.5755 (+0.2070) \\
 & DARE & 0.3438 (+0.1054) & 0.4652 (+0.1279) & 0.3623 (+0.1092) & 0.4856 (+0.1303) \\
 & DARE-TIES & 0.4529 (+0.1535) & 0.5624 (+0.1616) & 0.4880 (+0.1646) & 0.5899 (+0.1663) \\
 & Fisher & 0.4861 (+0.1165) & 0.6043 (+0.1051) & 0.5471 (+0.1368) & 0.6511 (+0.1160) \\
 & RegMean & 0.4880 (+0.1443) & 0.5959 (+0.1471) & 0.5250 (+0.1591) & 0.6295 (+0.1591) \\
\hline
\end{tabular}}
\caption{Submetric scores on the IFEval. Prompt-strict is the domain aggregate after applying \method{}, with the change against the anchor in parentheses.}
\label{tab:sub_instr}
\end{table*}

\begin{table*}[!t]
\centering{\small
\setlength{\tabcolsep}{4pt}
\begin{tabular}{llccccc}
\hline
Pool & Anchor & Coding macro & HE+ P@1 & MBPP+ P@1 & HE+ P@10 & MBPP+ P@10 \\
\hline
Llama-3.2-3B & Linear & 0.4343 (+0.0669) & 0.3838 (+0.0808) & 0.4847 (+0.0530) & 0.5350 (+0.0838) & 0.5985 (+0.0456) \\
 & TA & 0.4337 (+0.0644) & 0.3832 (+0.0610) & 0.4842 (+0.0679) & 0.5228 (+0.0567) & 0.5773 (+0.0723) \\
 & TIES & 0.4217 (+0.0296) & 0.3704 (+0.0234) & 0.4731 (+0.0358) & 0.5167 (+0.0106) & 0.5720 (+0.0323) \\
 & DARE & 0.4451 (+0.0632) & 0.3935 (+0.0585) & 0.4966 (+0.0678) & 0.5167 (+0.0384) & 0.5826 (+0.0644) \\
 & DARE-TIES & 0.4318 (+0.0485) & 0.3887 (+0.0607) & 0.4749 (+0.0363) & 0.5228 (+0.0594) & 0.5853 (+0.0430) \\
 & Fisher & 0.4309 (+0.0555) & 0.3893 (+0.0710) & 0.4726 (+0.0401) & 0.5106 (+0.0777) & 0.5800 (+0.0377) \\
 & RegMean & 0.4207 (+0.0481) & 0.3685 (+0.0551) & 0.4728 (+0.0411) & 0.4923 (+0.0350) & 0.5906 (+0.0324) \\
\hline
Llama-3.1-8B-Instruct & Linear & 0.6214 (+0.0403) & 0.6189 (+0.0461) & 0.6238 (+0.0344) & 0.7195 (+0.0583) & 0.6799 (+0.0321) \\
 & TA & 0.4941 (+0.2149) & 0.4568 (+0.4068) & 0.5313 (+0.0228) & 0.6154 (+0.5056) & 0.6226 (+0.0327) \\
 & TIES & 0.6064 (+0.0452) & 0.6012 (+0.0467) & 0.6116 (+0.0437) & 0.7073 (+0.0095) & 0.6720 (+0.0295) \\
 & DARE & 0.5754 (+0.1309) & 0.5779 (+0.2377) & 0.5729 (+0.0242) & 0.7273 (+0.2090) & 0.6523 (+0.0253) \\
 & DARE-TIES & 0.6151 (+0.0516) & 0.6201 (+0.0638) & 0.6101 (+0.0395) & 0.7561 (+0.0400) & 0.6852 (+0.0400) \\
 & Fisher & 0.6353 (+0.0444) & 0.6341 (+0.0528) & 0.6365 (+0.0360) & 0.6951 (+0.0156) & 0.7011 (+0.0427) \\
 & RegMean & 0.6263 (+0.0516) & 0.6220 (+0.0583) & 0.6307 (+0.0450) & 0.7256 (+0.0705) & 0.6878 (+0.0426) \\
\hline
Gemma-2-2B-it & Linear & 0.3958 (+0.0617) & 0.3659 (+0.0580) & 0.4257 (+0.0654) & 0.4390 (+0.0561) & 0.5132 (+0.0526) \\
 & TA & 0.3381 (+0.1196) & 0.3280 (+0.0926) & 0.3481 (+0.1465) & 0.4085 (+0.1158) & 0.4339 (+0.1138) \\
 & TIES & 0.3949 (+0.0604) & 0.3689 (+0.0415) & 0.4209 (+0.0794) & 0.4329 (+0.0256) & 0.5370 (+0.0791) \\
 & DARE & 0.3858 (+0.0859) & 0.3677 (+0.0726) & 0.4040 (+0.0992) & 0.4634 (+0.0683) & 0.5026 (+0.0579) \\
 & DARE-TIES & 0.3936 (+0.0632) & 0.3713 (+0.0530) & 0.4159 (+0.0733) & 0.4146 (+0.0317) & 0.5159 (+0.0685) \\
 & Fisher & 0.3917 (+0.0579) & 0.3500 (+0.0384) & 0.4333 (+0.0772) & 0.4024 (+0.0134) & 0.5159 (+0.0447) \\
 & RegMean & 0.3993 (+0.0654) & 0.3695 (+0.0591) & 0.4291 (+0.0717) & 0.4329 (+0.0622) & 0.5026 (+0.0500) \\
\hline
\end{tabular}}
\caption{Submetric scores on the coding domain after applying \method{}, with the change against the anchor in parentheses.}
\label{tab:sub_coding}
\end{table*}

\begin{table*}[!t]
\centering{\small
\setlength{\tabcolsep}{4pt}
\begin{tabular}{llcccc}
\hline
Pool & Anchor & Multi. macro & mMMLU & mARC & mHellaSwag \\
\hline
Llama-3.2-3B & Linear & 0.4817 (+0.0018) & 0.4738 (-0.0001) & 0.3920 (-0.0014) & 0.5793 (+0.0068) \\
 & TA & 0.4695 (-0.0083) & 0.4484 (-0.0100) & 0.3860 (-0.0095) & 0.5743 (-0.0050) \\
 & TIES & 0.4795 (-0.0001) & 0.4616 (-0.0057) & 0.3976 (+0.0046) & 0.5795 (+0.0011) \\
 & DARE & 0.4727 (-0.0053) & 0.4518 (-0.0081) & 0.3879 (-0.0059) & 0.5783 (-0.0020) \\
 & DARE-TIES & 0.4812 (+0.0017) & 0.4699 (-0.0016) & 0.3939 (+0.0020) & 0.5799 (+0.0048) \\
 & Fisher & 0.4833 (+0.0020) & 0.4723 (+0.0002) & 0.3971 (+0.0001) & 0.5804 (+0.0057) \\
 & RegMean & 0.4816 (+0.0034) & 0.4724 (-0.0003) & 0.3935 (+0.0033) & 0.5789 (+0.0073) \\
\hline
Llama-3.1-8B-Instruct & Linear & 0.5390 (+0.0168) & 0.5553 (+0.0141) & 0.4201 (+0.0168) & 0.6417 (+0.0197) \\
 & TA & 0.5022 (+0.0091) & 0.4995 (+0.0110) & 0.4010 (+0.0076) & 0.6061 (+0.0088) \\
 & TIES & 0.5202 (-0.0006) & 0.5422 (-0.0039) & 0.3915 (-0.0008) & 0.6270 (+0.0031) \\
 & DARE & 0.5232 (+0.0019) & 0.5291 (+0.0031) & 0.4120 (+0.0008) & 0.6285 (+0.0017) \\
 & DARE-TIES & 0.5284 (+0.0061) & 0.5475 (+0.0025) & 0.4050 (+0.0053) & 0.6329 (+0.0105) \\
 & Fisher & 0.5464 (+0.0069) & 0.5634 (+0.0026) & 0.4332 (+0.0122) & 0.6425 (+0.0057) \\
 & RegMean & 0.5391 (+0.0071) & 0.5553 (+0.0041) & 0.4201 (+0.0080) & 0.6419 (+0.0091) \\
\hline
Gemma-2-2B-it & Linear & 0.5012 (+0.0378) & 0.4981 (+0.0418) & 0.4224 (+0.0426) & 0.5832 (+0.0290) \\
 & TA & 0.4641 (+0.0501) & 0.4504 (+0.0700) & 0.3906 (+0.0338) & 0.5513 (+0.0465) \\
 & TIES & 0.5005 (+0.0435) & 0.4944 (+0.0560) & 0.4171 (+0.0401) & 0.5899 (+0.0344) \\
 & DARE & 0.4917 (+0.0491) & 0.4826 (+0.0597) & 0.4107 (+0.0414) & 0.5818 (+0.0463) \\
 & DARE-TIES & 0.5017 (+0.0412) & 0.4979 (+0.0506) & 0.4192 (+0.0419) & 0.5878 (+0.0310) \\
 & Fisher & 0.4980 (+0.0353) & 0.4955 (+0.0386) & 0.4152 (+0.0369) & 0.5832 (+0.0304) \\
 & RegMean & 0.5013 (+0.0372) & 0.4990 (+0.0420) & 0.4212 (+0.0405) & 0.5836 (+0.0291) \\
\hline
\end{tabular}}
\caption{Submetric scores on the multilingual domain (Okapi versions) after applying \method{}, with the change against the anchor in parentheses.}
\label{tab:sub_multi}
\end{table*}

\begin{table*}[!t]
\centering{\small
\setlength{\tabcolsep}{4pt}
\begin{tabular}{llccccc}
\hline
Pool & Anchor & Safety macro & WildGuard & HarmBench & XSTest & DAN \\
\hline
Llama-3.2-3B & Linear & 0.4540 (+0.0325) & 0.4930 (+0.0043) & 0.4463 (+0.0588) & 0.5883 (+0.0950) & 0.2883 (-0.0284) \\
 & TA & 0.5044 (+0.0872) & 0.5811 (+0.0977) & 0.4931 (+0.0987) & 0.6750 (+0.1039) & 0.2683 (+0.0483) \\
 & TIES & 0.4638 (+0.0255) & 0.5183 (+0.0230) & 0.4556 (+0.0243) & 0.6528 (+0.1061) & 0.2283 (-0.0517) \\
 & DARE & 0.5104 (+0.0640) & 0.5918 (+0.1111) & 0.4931 (+0.0550) & 0.7017 (+0.1184) & 0.2550 (-0.0283) \\
 & DARE-TIES & 0.4518 (+0.0209) & 0.5037 (+0.0364) & 0.4056 (+0.0025) & 0.6328 (+0.0928) & 0.2650 (-0.0483) \\
 & Fisher & 0.4437 (+0.0247) & 0.4983 (+0.0123) & 0.4244 (+0.0400) & 0.5972 (+0.1083) & 0.2550 (-0.0617) \\
 & RegMean & 0.4424 (+0.0386) & 0.4863 (+0.0270) & 0.4431 (+0.0806) & 0.5817 (+0.0550) & 0.2583 (-0.0084) \\
\hline
Llama-3.1-8B-Instruct & Linear & 0.8816 (+0.0791) & 0.8812 (+0.0574) & 0.8719 (+0.1069) & 0.9267 (+0.0489) & 0.8467 (+0.1034) \\
 & TA & 0.8172 (+0.0728) & 0.8584 (+0.0854) & 0.8181 (+0.0525) & 0.8889 (+0.0000) & 0.7033 (+0.1533) \\
 & TIES & 0.8596 (+0.0805) & 0.8785 (+0.0720) & 0.8500 (+0.0925) & 0.9267 (+0.0845) & 0.7833 (+0.0733) \\
 & DARE & 0.8651 (+0.0757) & 0.8758 (+0.0707) & 0.8369 (+0.0557) & 0.9244 (+0.0200) & 0.8233 (+0.1566) \\
 & DARE-TIES & 0.8953 (+0.1154) & 0.8838 (+0.0873) & 0.8875 (+0.1356) & 0.9400 (+0.0856) & 0.8700 (+0.1533) \\
 & Fisher & 0.9115 (+0.0599) & 0.9146 (+0.0527) & 0.8969 (+0.0857) & 0.9111 (+0.0544) & 0.9233 (+0.0466) \\
 & RegMean & 0.8730 (+0.0780) & 0.8919 (+0.0700) & 0.8469 (+0.0888) & 0.9200 (+0.0700) & 0.8333 (+0.0833) \\
\hline
Gemma-2-2B-it & Linear & 0.8927 (+0.1533) & 0.9465 (+0.2000) & 0.9656 (+0.1556) & 0.9067 (+0.0723) & 0.7520 (+0.1853) \\
 & TA & 0.7800 (+0.1615) & 0.8250 (+0.1494) & 0.7694 (+0.1600) & 0.9067 (+0.0311) & 0.6188 (+0.3055) \\
 & TIES & 0.8947 (+0.1641) & 0.9489 (+0.1891) & 0.9631 (+0.1962) & 0.9067 (+0.0645) & 0.7600 (+0.2067) \\
 & DARE & 0.8661 (+0.1852) & 0.8931 (+0.1640) & 0.9100 (+0.2244) & 0.9067 (+0.0511) & 0.7544 (+0.3011) \\
 & DARE-TIES & 0.8887 (+0.1637) & 0.9465 (+0.1760) & 0.9656 (+0.1894) & 0.9067 (+0.0600) & 0.7361 (+0.2294) \\
 & Fisher & 0.8637 (+0.1459) & 0.9345 (+0.1694) & 0.9656 (+0.1894) & 0.9067 (+0.0534) & 0.6482 (+0.1715) \\
 & RegMean & 0.8853 (+0.1530) & 0.9519 (+0.1734) & 0.9656 (+0.1737) & 0.9067 (+0.0578) & 0.7169 (+0.2069) \\
\hline
\end{tabular}}
\caption{Submetric scores on the safety domain after applying \method{}, with the change against the anchor in parentheses.}
\label{tab:sub_safety}
\end{table*}

\section{Limitations}
\label{app:limitations}
\method{} moves a checkpoint from a dense anchor toward the domain experts. It chooses neither of
those endpoints. The anchor comes from a procedure the repair does not run, and the experts are fixed
by the pool, so both ends of the interval the repair works inside are set before it starts.

\paragraph{The anchor fixes the level.}
Within each pool the anchor's Macro-5 and the post-repair Macro-5 correlate at $+0.96$, $+0.89$, and
$+0.86$, and 56 of the 63 within-pool anchor pairs keep their order across the repair. A strong anchor
stays strong and a weak one stays weak. The spread that single choice controls reaches 11.16 Macro-5
points between the weakest and the best anchor on Gemma-2-2B-it, wider than any gain the repair
produces. Because the ordering survives, taking the strongest dense merge gives up only 0.32 points
against choosing with hindsight, so the level is something a practitioner can steer. The size of the
gain is not. No anchor-side statistic of App.~\ref{app:anchorselect} predicts it on all three pools, which
leaves the anchor choice well informed about where the checkpoint will land and uninformed about how
far the repair will carry it.

\paragraph{The experts fix the ceiling.}
Eq.~\ref{eq:gap} clips each deficit at zero, so a domain the anchor already matches has $\rho_d=0$ and
is allocated no capacity at all. On Llama-3.2-3B all seven anchors match the multilingual expert.
Multilingual receives nothing there, and its score drifts by $-0.07$ points on average with whatever
the other four domains write, enough on two of the seven anchors to open gaps of 0.66 and 0.34 points
against that expert. The clip that makes the allocation follow measured need is the same clip that
leaves the repair nothing to aim at once a domain is level. \method{} recovers capability the anchor
gave away and does not reach past the pool it was built from.

\paragraph{Five domains is the range we measured.}
Every experiment here uses $|\mathcal D|=5$. The quota $C_i\rho_d n_i$ divides a fixed layer budget
among the domains sharing it, so it thins as the pool grows, and the one lever that would offset that
is the ceiling $C_{\max}$. Raising it is not free. Moving $C_{\max}$ from $0.45$ to $0.65$ costs 2.72 Macro-5 points on Llama-3.2-3B and 0.55
on Llama-3.1-8B-Instruct. Whether a larger pool is better served by a wider budget, by admitting
only the domains furthest behind, or by more than one round of occupancy is a question three pools of
five do not reach.

\clearpage
\clearpage
\section{Detailed Information about Reproducibility Checklist}
\label{app:release}
This section gives detailed information about each criterion in the reproducibility checklist.

\begin{itemize}
\item \textbf{1. General Paper Structure}
\item 1.1. Includes a conceptual outline and/or pseudocode description of AI methods introduced.\\
\textbf{[yes]}
\item 1.2. Clearly delineates statements that are opinions, hypothesis, and speculation from objective facts and results.\\
\textbf{[yes]}
\item 1.3. Provides well-marked pedagogical references for less-familiar readers to gain background necessary to replicate the paper.\\
\textbf{[yes]}
\item \textbf{2. Theoretical Contributions}
\item 2.1. Does this paper make theoretical contributions?\\
\textbf{[yes]} See Theorem~\ref{thm:occupancy} in App.~\ref{app:theory}.
\item 2.2. All assumptions and restrictions are stated clearly and formally.\\
\textbf{[yes]} See Assumption~\ref{ass:occupancy} and the Setting paragraph of App.~\ref{app:theory}, which fixes the layer, the quota rule, the precedence order, and the tie-breaking rule.
\item 2.3. All novel claims are stated formally (e.g., in theorem statements).\\
\textbf{[yes]} See Theorem~\ref{thm:occupancy}.
\item 2.4. Proofs of all novel claims are included.\\
\textbf{[yes]} See the proof in App.~\ref{app:theory}.
\item 2.5. Proof sketches or intuitions are given for complex and/or novel results.\\
\textbf{[yes]} See the proof intuition paragraph in App.~\ref{app:theory}.
\item 2.6. Appropriate citations to theoretical tools used are given.\\
\textbf{[yes]}
\item 2.7. All theoretical claims are demonstrated empirically to hold.\\
\textbf{[yes]} The theorem states that the sequential rule maximizes admitted mass at a fixed quota. The coordinate-selection control of App.~\ref{app:controls} holds that quota at its default value and claims coordinates at random instead, which retains 21.5, 21.2, and 22.0 percent of the paired gain.
\item 2.8. All experimental code used to eliminate or disprove claims is included.\\
\textbf{[yes]} The controls and ablations run through the same entry point as the default configuration, the \emph{run\_sigmerge} entry point, with the variant selected in its configuration file.
\item \textbf{3. Dataset Usage}
\item 3.1. Does this paper rely on one or more datasets?\\
\textbf{[yes]} Eleven benchmark variants, listed with their exact versions and record counts in Table~\ref{tab:split_membership}.
\item 3.2. A motivation is given for why the experiments are conducted on the selected datasets.\\
\textbf{[yes]} These eleven are the benchmarks that constitute the five MergeBench domains, so the expert pools and the evaluation suite match without reweighting. See App.~\ref{app:implementation}.
\item 3.3. All novel datasets introduced in this paper are included in a data appendix.\\
\textbf{[NA]} We introduce no novel dataset.
\item 3.4. All novel datasets introduced in this paper will be made publicly available upon publication of the paper with a license that allows free usage for research purposes.\\
\textbf{[NA]} We introduce no novel dataset.
\item 3.5. All datasets drawn from the existing literature (potentially including authors' own previously published work) are accompanied by appropriate citations.\\
\textbf{[yes]} Each benchmark is cited at first use in the main paper, and the exact variant of each is recorded in App.~\ref{app:implementation}.
\item 3.6. All datasets drawn from the existing literature (potentially including authors' own previously published work) are publicly available.\\
\textbf{[yes]} All eleven are public. The archive additionally ships the fixed partition we used, so the split does not have to be reconstructed.
\item 3.7. All datasets that are not publicly available are described in detail, with explanation why publicly available alternatives are not scientifically satisficing.\\
\textbf{[NA]} All datasets are publicly available.
\item \textbf{4. Computational Experiments}
\item 4.1. Does this paper include computational experiments?\\
\textbf{[yes]} See the Experiments section of the main paper and App.~\ref{app:results}.
\item 4.2. This paper states the number and range of values tried per (hyper-) parameter during development of the paper, along with the criterion used for selecting the final parameter setting.\\
\textbf{[yes]} Table~\ref{tab:hyperparameter_protocol} gives the three constants and the four values tried for each. The criterion is Macro-5 on Llama-3.2-3B, and the selected values are then transferred unchanged to the other two pools and all seven anchors. See App.~\ref{app:sensitivity}.
\item 4.3. Any code required for pre-processing data is included in the appendix.\\
\textbf{[yes]} See \emph{build\_splits} and \emph{verify\_splits} under \emph{data/scripts} in the archive. The delivered split holds 98,940 records, 19,784 of them calibration and 79,156 evaluation, and running the two verifiers in full mode against the shipped SHA-256 list reproduces those counts and confirms that the two partitions are disjoint.
\item 4.4. All source code required for conducting and analyzing the experiments is included in a code appendix.\\
\textbf{[yes]} See the \emph{sigmerge\_core} directory of the archive, whose six top-level entry points run the capacity profile, the domain deficits, the merge, the evaluation, the metric aggregation, and the paired statistics in that order.
\item 4.5. All source code required for conducting and analyzing the experiments will be made publicly available upon publication of the paper with a license that allows free usage for research purposes.\\
\textbf{[yes]} We will release the code publicly upon publication.
\item 4.6. All source code implementing new methods have comments detailing the implementation, with references to the paper where each step comes from.\\
\textbf{[yes]} See the package under \emph{sigmerge\_core}, where the conflict,
deficit, and occupancy modules follow the same order as the Method section of the main paper.
\item 4.7. If an algorithm depends on randomness, then the method used for setting seeds is described in a way sufficient to allow replication of results.\\
\textbf{[yes]} Seed values are $\{0,1,2\}$ for the two stochastic anchors and MERGE$^3$ and
$\{0,\ldots,4\}$ for the all-random control. The benchmark partition and generation-based evaluation
each use a fixed seed. See App.~\ref{app:implementation} and App.~\ref{app:perseed}.
\item 4.8. This paper specifies the computing infrastructure used for running experiments (hardware and software), including GPU/CPU models; amount of memory; operating system; names and versions of relevant software libraries and frameworks.\\
\textbf{[yes]} See App.~\ref{app:implementation}. The evaluation launcher additionally expects local checkouts of three upstream evaluators at the commits we ran, lm-evaluation-harness at ee7e8f4fe58e, bigcode-evaluation-harness at 8fc5bae6479c, and safety-eval at 2920bb85a8a8, whose full hashes the archive records.
\item 4.9. This paper formally describes evaluation metrics used and explains the motivation for choosing these metrics.\\
\textbf{[yes]} See the metric construction paragraph of App.~\ref{app:implementation} and the submetric definitions of App.~\ref{app:submetrics}.
\item 4.10. This paper states the number of algorithm runs used to compute each reported result.\\
\textbf{[yes]} \method{} is deterministic given a reference, an expert pool, an anchor, and a calibration split, so each of its rows is one run. The two stochastic anchors and the search-based baseline use three seeds and the all-random control uses five. See App.~\ref{app:perseed}.
\item 4.11. Analysis of experiments goes beyond single-dimensional summaries of performance (e.g., average; median) to include measures of variation, confidence, or other distributional information.\\
\textbf{[yes]} App.~\ref{app:perseed} reports per-seed values with standard deviation and range, App.~\ref{app:shares} breaks every aggregate out by domain, and App.~\ref{app:submetrics} goes one level further to the seventeen submetrics.
\item 4.12. The significance of any improvement or decrease in performance is judged using appropriate statistical tests (e.g., Wilcoxon signed-rank).\\
\textbf{[yes]} A two-sided exact Wilcoxon signed-rank test over the 21 anchor-pool pairs gives $p=9.5\times10^{-7}$, and the deficit-gain relation is reported as a Spearman correlation over the 105 triples. Both are recomputed by the \emph{analyze\_results} entry point of the archive.
\item 4.13. This paper lists all final (hyper-)parameters used for each model/algorithm in the paper's experiments.\\
\textbf{[yes]} See Table~\ref{tab:hyperparameter_protocol} for \method{}. Every baseline runs with the constants its own paper reports, as recorded in the caption of Table~\ref{tab:efficiency}.
\end{itemize}